\documentclass[12pt]{article}

\usepackage{amssymb,amsmath,amsthm,amsfonts}
\usepackage[all,arc]{xy}
\usepackage{enumerate}
\usepackage{mathrsfs}
\usepackage{hyperref}
\usepackage[titletoc]{appendix}

\usepackage{tikz}
\usetikzlibrary{arrows, positioning}
\usetikzlibrary{automata,arrows,positioning,calc,chains,patterns}
\usepackage{graphicx}
\usepackage{subcaption}

\usepackage{empheq}
\newcommand*\widefbox[1]{\fbox{\hspace{0.5em}#1\hspace{0.5em}}}

\makeatletter
\let\c@equation\c@thm
\makeatother
\numberwithin{equation}{section}

\newcommand{\ba}{\mathbf a}

\newcommand{\bo}{\mathbf o}
\newcommand{\bs}{\mathbf s}

\newcommand{\bx}{\mathbf x}
\newcommand{\by}{\mathbf y}
\newcommand{\bM}{\mathbf M}
\newcommand{\bp}{\mathbf p}
\newcommand{\bq}{\mathbf q}
\newcommand{\bA}{\mathbf A}
\newcommand{\bB}{\mathbf B}
\newcommand{\bD}{\mathbf D}

\newcommand{\bOne}{\mathbf 1}

\newcommand{\bE}{\begin{equation}}
\newcommand{\eE}{\end{equation}}
\newcommand{\bBl}{\begin{align}}
\newcommand{\eAl}{\end{align}}

\newcommand{\bpi}{\boldsymbol \pi}
\newcommand{\bmu}{\boldsymbol \mu}
\newcommand{\balpha}{\boldsymbol \alpha}
\newcommand{\bbeta}{\boldsymbol \beta}
\newcommand{\bsigma}{\boldsymbol \sigma}

\newcommand{\KL}[2]{{\mathrm{KL}\,}\!\left[#1\,\|\, #2 \right]}
\newcommand{\E}[1]{{\mathbb {E}\,}\!\left[#1\right]}
\newcommand{\Esub}[2]{{\mathbb {E}_{#1}\,}\!\left[#2\right]}
\newcommand{\ent}[1]{{\mathrm {H}\,}\!\left[#1\right]}
\newcommand{\F}[1]{{\mathrm {F}\,}\!\left[#1\right]}

\title{Active Inference in Discrete State Spaces from First Principles}

\author{Patrick Kenny \\ {\em Patrick.Kenny.000@gmail.com} \\ {\em Centre de recherche informatique de Montr\'eal}
}

\begin{document}

\tikzset{
modal/.style={>=stealth',shorten >=1pt,shorten <=1pt,auto,
node distance=1.0cm,semithick},
world/.style={circle,draw,minimum size=1.5cm, pattern=north east lines, pattern color=gray},
point/.style={circle,draw,minimum size=1.5cm}
}

\maketitle

\begin{abstract} 
We seek to clarify the concept of active inference by disentangling it from the Free Energy Principle. We show how the 
optimizations that need to be carried out in order to implement active inference in discrete state spaces can be formulated as constrained divergence minimization problems which can be solved by standard mean field methods that do not appeal to the idea of expected free energy.  When it is used to model perception, the perception/action divergence criterion that we propose coincides with variational free energy. When it is used to model action, it 
differs from an expected free energy functional by an entropy regularizer.
\end{abstract}

\tableofcontents

\section{Introduction}

The idea that sense perception can be understood as an unconscious process of inference dates back to 
Helmholtz. Among his many other scientific contributions is
the concept of Helmholtz free energy which quantifies the energy in a system that is available to do physical work such as moving a piston. Hinton and his collaborators 
realized that approximate Bayesian inference in machine learning could be viewed as a problem of optimizing an abstract mathematical quantity which is formally
identical to Helmholtz free energy and which has come to be known as variational free energy. (Thus the Helmholtz machine \cite{Dayan1995Helmholtz} 
and the Boltzmann machine \cite{Salakhutdinov2009DBM}.) In a long series of papers (notably \cite{Friston2003,Friston2005,Friston2010}), Friston has developed the connections
between these ideas and greatly expanded their scope by showing how action as well as sense perception can be modelled as approximate Bayesian inference.

If perceptions are inferences drawn from bodily sensations, then actions can be viewed as fulfilling predictions of future sensations. Whereas
perceptual inference is thought to consist of computing approximate posterior probability distributions, active inference is a matter of computing approximate predictive distributions. 
Both types of inference can be formulated as optimization problems which are amenable to the sort of variational inference algorithms that have been developed in machine learning.
In the case of perception, the objective function to be optimized is just variational free energy and this optimization can be achieved by the well known 
mean field approximation \cite{Blei2017VariationalIR}.
In the case of action, a new type of objective function known 
as expected free energy has been proposed and new variational methods have been developed to optimize it
\cite{oostrum_concise_2025, Smith2022, DaCosta2020_DiscreteActiveInference,Sajid2021Active, zhang2024_overview_fep}.

``It is said that the Free Energy Principle is difficult to understand."  Thus the opening sentence of a review article that purports to simplify the Free Energy Principle 
but brooks no compromise on Friston's program of grounding variational inference by biological agents in statistical physics \cite {FristonEtAl2023}. For a physicist studying
active inference, it is natural to assume that a biological agent models its world as a random
dynamical system governed by a stochastic differential equation and to construe the agent's striving to maintain homeostasis as a pullback attractor. However, in simulating the behaviour of  biological agents or in designing
autonomous AI agents capable of multistep hierarchical planning, an engineer would assume instead that an agent models its world as a discrete state space that evolves in discrete time steps using a Hidden Markov Model or 
Partially Observable Markov Decision Process.  The mathematical apparatus that the engineer needs to deploy 
is much simpler than that required by the physicist. 

In this paper we aim to give a self-contained and mathematically rigorous account of active inference in discrete state spaces without appealing to any of the machinery 
that has been developed in the context of continuous state spaces. One advantage of the discrete set up is that Hidden Markov Models accommodate the path integral formulation of the active inference problem quite readily.  Another is that mixed continuous/discrete state models can be avoided since it is natural to model the actions available to an agent in a given situation as a discrete set and the actions themselves as discrete transition probability matrices. This enables the distinction between states and actions to be dissolved by augmenting the definition of a state to include a tag which indicates which action is currently underway. So a sufficiently rich Hidden Markov Model structure obviates the need to refer to actions explicitly. Sequences of actions (and even sequences of sequences of actions) can be handled in the same way as actions since they too can be modelled by transition probability matrices \cite{friston_pixels_2024}.  Thus we can conceive of a type of agent that is endowed with a Hidden Markov Model whose structure is sufficiently rich that it can encode {\it all} of the agent's knowledge of the dynamics of the world it inhabits, including its own actions and policies and their sensory consequences.  For such an agent, perceptual and active inference would consist in using this Hidden Markov Model to calculate posterior and predictive probability distributions conditioned on its history up to the present moment, if only these computations were tractable. Given that the agent does not have the resources to calculate these distributions exactly, it needs to resort to methods of calculating approximate posterior and predictive distributions that yield results which are accurate enough to be actionable. 

If active inference is viewed in this way, there is no {\em a priori} requirement to appeal to the Free Energy Principle.  So, although active inference and the Free Energy Principle are usually conflated (as in the title of the active inference textbook \cite{Parr2022ActiveInference}, for example), we will distinguish between the two. As we will see, active inference in discrete state spaces (understood as the problem of inferring predictive probability distributions for an agent's future sensations) is amenable to treatment by 
standard mean field methods which do not appeal to the idea of expected free energy. We will show how this approach enables us to model the  perception/action cycle in a unified way by optimizing a single Kullback-Leibler divergence criterion rather than by optimizing variational free energy to model perception and expected free energy to model action, as is usually done. We will also show how the mean field approximation can be brought to bear on other aspects of the active inference problem, namely Bayesian learning of Hidden Markov Model parameters and updating beliefs about policies.

We will briefly review the Bayesian Brain Hypothesis, and basic ideas in Bayesian machine learning and statistical mechanics before getting down to business. Readers who have had no previous exposure to the subject of active inference may find
it useful to consult popularizations by Andy Clark \cite{clark2015_surfinguncertainty}, Anil Seth \cite{seth2021_beingyou} and Mark Solms \cite{solms2021_hiddenspring}.

\section{The Bayesian Brain}

In this section,  we sketch the Bayesian Brain Hypothesis, Active Inference and the Free Energy Principle and we lay out a roadmap for the paper.

\subsection{Probabilities and Beliefs}

\begin{quote}
In the greatest part of our concernments, [God] has afforded us only the twilight, as I may so say, of probability; suitable, I presume, to that state of mediocrity and probationership he has been pleased to place us in here; wherein, to check our over-confidence and presumption, we might, by every day's experience, be made sensible of our short-sightedness and liableness to error.
\end{quote}

Thus John Locke,  writing in 1689 \cite{Locke1689-LOCAEC-4}. What Locke understood by probability is unclear and, although the mathematical theory of probability has long been well established, there is still no consensus among statisticians, engineers and physicists as to what probabilities mean. Neuroscientists however have settled on the Bayesian interpretation of probabilities as credences or degrees of belief. (This explains why the terms ``probability distribution" and 
``belief" have come to be used interchangeably in the active inference literature.)

Under this interpretation, probabilities refer to the state of an observer's knowledge of events rather than to objective knowledge of the events themselves. Physical scientists generally balk at this interpretation 
for obvious reasons, but it has the great advantage of formalizing the problem of reasoning under uncertainty in a very parsimonious way: all inference, without exception, reduces to the sum and product rules for combining probabilities.  
Provided that numerical values are correctly assigned to beliefs and probability calculations are carried out exactly, Bayesian decision theory is provably optimal. 

On the Bayesian Brain Hypothesis, the brain is a repository of beliefs about how the body and environment work and how they interact with each other. 
Bayesian beliefs about events are subject to revision as more information becomes available. (More formally: prior probabilities are converted to posterior probabilities by applying Bayes’ rule.)  
As the brain collects sense data from the body in a given situation, it infers the causes of its sensations --- in other words, what is going on in its world --- by approximate Bayesian inference 
which converts its prior beliefs (about how the world works in general) into posterior beliefs (about what the world is doing right now). These posterior beliefs are what we think of as reality.

\subsection{Perception as Probabilistic Inference}
\label{sec:perception}

As a first step towards a general formulation of the Bayesian Brain Hypothesis, consider that the primary function of the brain is to regulate the body and to act on the environment in such a way as to maintain the body in a state of homeostasis. All of the brain's knowledge of what is going 
on in its world (which includes the body as well as the environment) at a given time is mediated by incoming sensory data. We can model this state of affairs with a steady state joint probability distribution $p(\bs, \bo|S)$ where $\bs$ is a vector whose components represent the state of the organism's world and  $\bo$ is a vector comprised of all of the incoming sensory data (interoceptive and proprioceptive as well as exteroceptive). The state vector $\bs$ can be thought of as comprising the causes of the observation vector $\bo$.
The marginal distribution $p(\bs|S)$ encodes the brain's prior knowledge of its world. The state of the world at a given time is hidden in the sense that it cannot be directly observed but has to be inferred from incoming sense data.  From this perspective, sense perception is just this process of inference. 

The terminology ``observation" should not be read as suggesting that sense data are generally accessible to consciousness.  We are normally not consciously aware of most of our visceral and proprioceptive sensations, 
just as we are unaware of the sensations of sound pressure waves impinging on our ears (except in the case of loud noises) or of light waves impinging on our eyes (except in the case of sudden flashes). 
Consciousness appears mysterious to us because, along with many of the sensations themselves, 
the process which converts sensations to perceptions is hidden from us. 

For a given observation vector $\bo$, the marginal probability $p(\bo|S)$ is known as the {\em model evidence} and its negative logarithm, $-\! \ln p(\bo|S)$, as the {\em surprisal}. (The less probable an event, the more surprising it is.)
In principle the evidence or surprisal could be evaluated by summing over all possible states of the world:
\[
p(\bo|S) = \sum_s p(\bs, \bo|S)
\]
but this summation is obviously intractable. However if the evidence could be evaluated, the
problem of inferring the state of the world from sensory data could be solved by a direct calculation of the posterior probability $p(\bs|\bo, S)$:
\bE
p(\bs|\bo, S) = \frac{p(\bs, \bo|S)}{p(\bo|S)}.
\eE

Approximations cannot be avoided and, given the scale of the problem, it is reasonable to assume that the brain uses
variational approximations like those that have been developed in machine learning rather than Monte Carlo methods.
So we postulate that the brain calculates some sort of mean field approximation to the posterior distribution $p(\bs|\bo, S)$. 
As a byproduct, this calculation gives an approximation to the model evidence which is known in machine learning as the evidence lower bound or ELBO.
{\em Variational free energy} is defined by changing the sign
on the ELBO, so that one talks about minimizing the variational free energy rather than maximizing the ELBO and minimizing the surprisal 
rather than maximizing the evidence.

On this account, variational free energy is constantly being minimized as 
the approximate posterior distribution is updated from one moment
to the next. It is proposed in \cite{Hobson2014-HOBCDA} that qualia can be understood as transductions of bodily sensations which are encoded in the brain by the 
parameters which specify this approximate posterior 
distribution in much the same way as photographs are encoded in jpeg files. (By a happy coincidence, 
variational posterior distributions in machine learning are traditionally denoted by $q$ rather than $p$ which is reserved for priors and exact posteriors.)

Note that approximating the full posterior distribution $p(\bs|\bo, S)$ 
(rather than merely computing a point estimate of $\bs$) enables the brain to assess the uncertainty attaching to its explanation of the sense data $\bo$. 
This is important because an agent cannot act effectively on its environment 
if it is uncertain about the state of its world at a given time or in its immediate future. In such a situation,  gathering information with a view to reducing uncertainty is a primary imperative.

\subsection{Active Inference and the Free Energy Principle}

In order to extend the Bayesian Brain Hypothesis to account for action as well as perception, Friston observes that  biological agents always seem to act in ways
that can be interpreted as striving to maximize the evidence for their model of the world or, since the model evidence is intractable, to minimize variational free energy \cite{Friston2013}.
This behaviour is often referred to as  ``self-evidencing'' in the active inference literature. Thus the agent is said to act on its world in such a way as 
to maximize the evidence for its continued existence or to minimize its surprise at what happens to it.

To this end, a biological agent needs to be able to predict its future sensations in much the same way as
large language models predict text. Whereas the predictions made by large language models are driven by a past history and the statistics of natural languages, the
predictions made by a biological agent are driven by its history up to the present moment and the requirement 
that future sensations are drawn from the steady state marginal distribution $p(\bo|S)$. Action is then construed as fulfilling these predictions. 
More specifically, actions ultimately reduce to autonomic and motor reflexes (that is, secreting hormones and contracting muscles) which fulfill predictions of 
interoceptive and proprioceptive sensations \cite{Adams2012PredictionsNC}.  This move brings action under the umbrella of the Bayesian Brain Hypothesis without invoking 
extrinsic ideas about goals, rewards or utility functions. Action is governed by prediction, that is, by probabilistic inference of the future states of the world.
Hence the term {\em active inference}. 

Predictions in exteroceptive sense modalities such as seeing and hearing over which the agent has limited control may be less
precise, but navigating an environment obviously requires the ability to anticipate events and not merely to react to events after the fact. Indeed, it is 
difficult to see how a biological agent could learn to act effectively on the world it finds itself inhabiting other than by learning 
to predict the sensory consequences of its behaviour in much the same way as large language models learn the statistics of natural languages
by learning to predict text one word at a time.

Predicting the future requires a probability model which is capable of tracking trajectories in the state space over time. (The steady state distribution
$p(\bs, \bo|S)$ that we have been considering is not adequate in this respect since it treats states at successive times as being statistically independent.)
The obvious choice is to model state trajectories as a Markov chain by assigning to each state a probability distribution over states that can be visited next.
This type of model is usually referred to as a Hidden Markov Model (HMM) rather than a Markov chain since the state occupied at a given time cannot be directly observed 
but has to be inferred probabilistically. If the states were 
observable, trajectories in the state space would appear to be directed towards regions that have high probability 
under the steady state marginal distribution $p(\bs|S)$, subject to perturbations due to random shocks coming from the agent's environment. 

A HMM is a joint distribution on
sequences of states and observations, rather than on individual states and observations as in the case of the steady state distribution.
Under this sort of dynamic model, perception is understood to be a matter of recognizing patterns that unfold over time 
(rather than static objects) and variational free
energies are associated with sequences of observations (rather than with individual observations).
So an agent contemplating an action could assign a variational free energy to it if it could foretell how the sensory consequences of the action would unfold.
Of course the agent cannot foresee the future but, for each action that is possible in a given situation, it 
could calculate an expected value for the variational free energy of the sense data it would encounter if the action was performed.
This would enable the agent to assign approximate probabilities to the various possible actions that are open it at a given time 
and use these probabilities in inferring a predictive distribution for its future sensations by variational methods.
The {\em Free Energy Principle} asserts that biological agents ``decide'' on courses of action in this way. 
(On Friston's account \cite{FristonAeon}, the ability to look ahead in performing this sort of variational inference is a hallmark of conscious behaviour.)

On the other hand, we will show in this paper that  the problem of active inference, understood as the problem of inferring a 
probability distribution for an agent's future sensations, can be solved by standard mean field methods that do not appeal to the idea of expected free energy.

\subsection{Road Map}

The Free Energy Principle suggests how an objective function known as expected free energy might be defined which would enable 
active inference to be cast as an optimization problem in the same way as perceptual inference. 
But, because it purports to be a general principle like Hamilton's principle of stationary action, the Free Energy Principle does not commit to a precise definition of this objective function. 
In this paper, we will proceed from the assumption that, for an agent that uses a HMM to model its world, the Kullback-Leibler divergence of a predictive distribution from the HMM is a natural measure of the quality of the predictive distribution. 
So rather than appealing to the Free Energy Principle, we posit that a (suitably constrained) Kullback-Leibler divergence can serve as the objective function for active inference 
and we show how it can be optimized by standard mean field methods.

The mean field approximation is generally used to infer approximate posterior distributions of hidden variables after data has been collected.
Thus, given a HMM $p(\bs, \bo)$ and a sequence of observations up to time $t$ denoted by $\bo_{\le t}$, minimizing variational free energy provides an approximation to the posterior distribution $p(\bs_{\le t}|\bo_{\le t})$ where $\bs_{\le t}$ is the hidden state sequence that accounts for $\bo_{\le t}$.
(The lower bound on the model evidence $p(\bo_{\le t})$ known as the ELBO is a byproduct of this calculation.) We will explain the mean field approximation in 
Sections \ref{sec:ML} and \ref{sec:StatMech}. We will show how it can be used to provide an approximation to the predictive distribution $p(\bs_{>  t}, \bo_{> t}|\bo_{\le t})$ on future states and observations  in addition to the posterior 
distribution  $p(\bs_{\le t}|\bo_{\le t})$ in Section \ref{sec:P/A}.  The core idea is very simple: it consists in treating future states and observations $\bs_{> t}$ and  $\bo_{> t}$ as hidden variables on the same footing as $\bs_{\le t}$ 
and calculating a variational posterior (conditioned on the observations $\bo_{\le t}$) for this augmented set of hidden variables. 
This perspective enables us to treat the perception/action cycle in a unified way.  We model it by minimizing a single Kullback-Leibler divergence functional rather 
than minimizing variational free energy in the case of perception and expected free energy in the case of action, as is usually done.

It is shown in \cite{hafner_action_2022} how a wide variety of reinforcement learning algorithms can be formulated as solving constrained divergence minimization problems and the 
exploration/exploitation tradeoff in reinforcement learning is analyzed in this general setting. We discuss how this applies to HMMs in Section \ref{sec:info}.  

Section \ref{sec:learning} shows how mean field methods apply to the problem of learning HMM parameters. Although this material is not new, we include it for completeness.

We discuss the relationship between our perception/action divergence criterion and various expected free energy functionals in Section \ref{sec:whither}. 
We explain the (implicit or explicit) approximations that are used to derive these expected free energy functionals and how the divergence criterion avoids them. 
It turns out that, when it is used to calculate predictive distributions, the perception/action divergence criterion only differs from a free energy functional by an entropy regularizer, although
the functional in question is not one that has been used in the active inference literature.

\section{Bayesian Machine Learning}
\label{sec:ML}

This section explains the mean field approximation and the concept of variational free energy. We begin with an overview of the rudiments of information theory and Bayesian machine learning. 

\subsection{Entropy, Cross Entropy and Divergence}
\label{sec:entropy}

We use the notation $p(\bx)$  to refer to a generic probability distribution of a vector valued random variable $\bx$ and $q(\bx)$  to refer to an approximate probability distribution. Typically, the distribution $p(\bx)$ is computationally intractable and $q(\bx)$ is constrained to belong to a family of tractable distributions.  For the most part, we take the values of  $\bx$ to be discrete. 

In situations where we need to distinguish between a random variable $\bx$ and a value it takes, we denote such a value by $\bar \bx$. For example, we use the notation $\delta_{\bar \bx}(\bx)$ to refer to the probability distribution all of whose mass is concentrated on the value $\bar \bx$:
\[
\delta_{\bar \bx}(\bx) = \begin{cases}
1 & \text{if $\bx = \bar \bx$} \\
0 & \text{otherwise.}
\end{cases}
\]

The {\em information content} or {\it surprisal} of the event that the random variable takes the value $\bar \bx$  is defined to be the negative log probability, $-\ln p(\bar \bx)$. (The more unlikely an event, the greater the surprise at its occurrence.)
The {\em entropy} of $p(\bx)$ and the {\em cross entropy} of $p(\bx)$ relative to $q(\bx)$ are defined by averaging the information with respect to $p(\bx)$ and $q(\bx)$ respectively:
\begin{align*}
\Esub{p(\bx)}{-\ln p(\bx)} &= -\sum_\bx p(\bx) \ln p(\bx) \\
\Esub{q(\bx)}{-\ln p(\bx)} &=  -\sum_\bx q(\bx) \ln p(\bx).
\end{align*}
We denote the entropy of $p(\bx)$  by $\ent{p(\bx)}$. This has the property that 
$$0 \le \ent{p(\bx)} \le \ln K$$ 
where $K$ is the number of values that $\bx$ can take. The minimum value of 0 is attained by point mass distributions and the maximum value by the uniform distribution which assigns equal probabilities to all values of $\bx$. 
So $\ent{p(\bx)}$ can be thought of as a measure of how flat the distribution $p(\bx)$ is.

The Kullback-Leibler divergence of $q(\bx)$ from $p(\bx)$ is the difference between the cross entropy and the entropy:
\begin{align}
\KL{q(\bx)}{p(\bx)} &= \Esub{q(\bx)}{-\ln p(\bx)} - \ent{q(\bx)} \label{eqn:KLCE} \\
&= \sum_\bx q(\bx) \ln \frac{q(\bx)}{p(\bx)}. \nonumber
\end{align}
It has the property that $\KL{q(\bx)}{p(\bx)} \ge 0$ with equality holding iff $q(\bx)$ and $p(\bx)$ are identical. Note that the divergence is not symmetric in its arguments (for more on this see \cite{murphy_probabilistic_2023}). 

Either the {\em forward divergence} $\KL{p(\bx)}{q(\bx)}$ (the ``divergence of $p(\bx)$ from $q(\bx)$") or the {\em reverse divergence} $\KL{q(\bx)}{p(\bx)}$ (the ``divergence of $q(\bx)$ from $p(\bx)$") can be used
to evaluate how well the target distribution $p(\bx)$ is approximated by $q(\bx)$.  We will mostly use reverse divergences in this paper as we will usually be dealing with
situations where the target distribution $p(\bx)$ is computationally intractable and the approximating distribution $q(\bx)$ is tractable so that the reverse divergence can be evaluated and optimized whereas the forward divergence cannot.

A situation where it is natural to use the forward divergence as the optimization criterion arises when the target distribution $p(\bx)$
is given empirically in the form of a training data set.  Since the entropy $\ent{p(\bx)}$ is independent of $q(\bx)$, minimizing the forward divergence $\KL{p(\bx)}{q(\bx)}$ 
is equivalent to minimizing the cross entropy $\Esub{p(\bx)}{- \ln q(\bx)}$. This is generally referred to as the cross-entropy loss function in non-Bayesian machine learning and minimizing it
 is equivalent to maximum likelihood estimation of $q(\bx)$ \cite{murphy_probabilistic_2023}. It is well known that maximum likelihood estimation is vulnerable to overfitting. On the other hand, when the reverse divergence is used as the optimization criterion, the 
the entropy $\ent{q(\bx)}$ which appears with a negative sign in (\ref{eqn:KLCE}) mitigates against this tendency by penalizing approximating distributions of low entropy.  It is often referred to as an ``entropy regularizer'' for this reason.
 
Reverse divergences arise naturally when posterior distributions need to be approximated.
Suppose we seek to minimize the divergence $\KL{q(\bx)}{p(\bx)}$ subject to the constraint that $q(\bx) = 0$ outside of a set $A$ so that the optimand is
\[
\sum_{\bx \in A} q(\bx) \ln \frac{q(\bx)}{p(\bx)}.
\] 
Define a probability distribution $p'(\bx)$ by
\bE
\label{eqn:000}
p'(\bx) = \begin{cases}
\frac{1}{p(A)}p(\bx) & (\bx \in A) \\
0 & (\bx \notin A)
\end{cases}
\eE
so that
\begin{align*}
\sum_{\bx \in A} q(\bx) \ln \frac{q(\bx)}{p(\bx)} &= \sum_{\bx \in A} q(\bx) \ln \frac{q(\bx)}{p'(\bx)} - \ln p(A).
\end{align*}
The first term on the right-hand side is just the divergence of $q(\bx)$ from $p'(\bx)$ 
and we can ignore the second term since it is independent of $q(\bx)$. If no additional constraints are imposed on $q(\bx)$, then the divergence assumes the minimum value of 0 when $q(\bx) = p'(\bx)$. 
Since the expression in the first line of (\ref{eqn:000}) is just the conditional distribution of $\bx$ given that $\bx \in A$, 
what this calculation shows is that Bayes rule can be derived by solving a constrained divergence minimization problem. All of the calculations that we carry out in this paper will follow this pattern.

We will frequently need to calculate the divergence of one joint distribution from another. For this we will use the following ``chain rule".
Given two joint distributions $q(\bx, \by)$ and $p(\bx, \by)$, the divergence  $\KL{q(\bx, \by)}{p(\bx, \by)}$ can be written in either of the forms
\setlength\fboxsep{0.1cm}
\begin{empheq}[box=\widefbox]{align}
& \KL{q(\bx)}{p(\bx)}   + \Esub{q(\bx)}{\KL{q(\by|\bx)}{p(\by|\bx)}} \nonumber  \\
& \KL{q(\by)}{p(\by)}  + \Esub{q(\by)}{\KL{q(\bx|\by)}{p(\bx|\by)}}  \label{eqn:chain}
\end{empheq}
This is a straightforward consequence of the definition of the divergence.

\subsection{The Mean Field Approximation}
\label{sec:MFA}

We can associate an {\em energy function} $E(\bx)$ with a probability distribution $p(\bx)$ by setting
$
E(\bx) = - \ln p(\bx)
$
so that 
$
p(\bx) = \exp\left(-E(\bx)\right).
$
(In physics, low energy states are more probable than high energy states. Hence the negative sign.)
Conversely, given an energy function $E(\bx)$, we can define a probability distribution $p(\bx)$ by setting
\[
p(\bx) = \frac{1}{Z} e^{-E(\bx)}
\]
where the {\em partition function} $Z$ is determined by the requirement that probabilities sum to 1. 

A well known example is the 
multivariate Gaussian distribution which is defined by an energy function of the form
$\frac{1}{2}\bx^\top {\boldsymbol \Sigma}^{-1} \bx$. 
In a situation where the state space is of high dimension and sparsity constraints are
imposed on the precision matrix  ${\boldsymbol \Sigma}^{-1}$, this distribution is referred to as a  { Gaussian Markov Random Field}.
If $\bx$ is constrained to be binary valued, the corresponding distribution is
known as the Ising model in statistical physics and as the Boltzmann machine in machine learning.  All of these distributions
are intractable in the case of high dimensional state spaces in the sense that the partition function cannot be evaluated.

Regardless of whether $p(\bx)$ is tractable or not, we can seek to approximate it by a tractable distribution $q(\bx)$
using the reverse divergence $\KL{q(\bx)}{p(\bx)}$ as the optimization criterion. In the mean field approximation, $\bx$ 
is decomposed into subvectors $\bx_0, \ldots, \bx_N$ and statistical independence constraints are imposed so that the approximating distribution $q(\bx)$ factorizes as
\bE
\label{eqn:17}
q(\bx_0)q(\bx_1) \ldots q(\bx_N).
\eE
The procedure that we will present for calculating $q(\bx)$ (which is referred to as coordinate ascent variational inference in 
\cite {Blei2017VariationalIR}), requires that  the factors are updated  {\em asynchronously} (that is, one at a time rather than in parallel). 
The divergence $\KL{q(\bx)}{p(\bx)}$ is guaranteed to decrease on each update but,
although the procedure is guaranteed to converge, it may converge to a local rather than a global minimum of the divergence. So it
 may need to be initialized carefully.

The rule for updating a factor $q(\bx_n)$ holding the remaining factors fixed is very simple: $q(\bx_n)$ is the distribution defined
by the energy function $\Esub{q(\bx_{\backslash n})} {E(\bx)}$ where $\bx_{\backslash n}$ denotes the set of components of $\bx$
other than $\bx_n$. To see that this update is guaranteed to decrease the divergence  $\KL{q(\bx)}{p(\bx)}$ let us write the divergence in the form
\[
\Esub{q(\bx_n)} {
\Esub{q(\bx_{\backslash n})} {\ln q(\bx) - \ln p(\bx)}
}.
\]
We need to minimize this with respect to $q(\bx_n)$. Using the notation $+ \ldots$ to indicate terms which are independent of $\bx_n$ and hence irrelevant,
\[
\ln q(\bx) = \ln q(\bx_n) + \ldots,
\]
so minimizing the divergence is equivalent to minimizing 
\bE
\label{eqn:MF}
\Esub{q(\bx_n)} {\ln q(\bx_n) - \ln \tilde q(\bx_n)}
\eE
where $\tilde q(\bx_n)$ is defined by
\[
\ln \tilde q(\bx_n) = \Esub{q(\bx_{\backslash n})} {\ln p(\bx)}.
\]
Generally speaking, the vector  $\tilde q(\bx_n)$ does not define a probability distribution because its components may not sum to 1, but if they did the expression (\ref{eqn:MF}) would be the divergence from $q(\bx_n)$ to $\tilde q(\bx_n)$ 
and it could be minimized by setting $q(\bx_n) = \tilde q(\bx_n)$. Since rescaling the vector $\tilde q(\bx_n)$ only changes (\ref{eqn:MF})
by an irrelevant additive constant, it follows that the update formula for $q(\bx_n)$ is $q(\bx_n) \propto \tilde q(\bx_n)$ where the constant of proportionality is determined by the condition that the components of $q(\bx_n)$ sum to 1.  
In sum:
\setlength\fboxsep{0.1cm}
\begin{empheq}[box=\widefbox]{align}
\label{eqn:VBupdate}
q(\bx_n) \propto \tilde q(\bx_n) \text{\hspace{0.0cm} where \hspace{0.0cm}} \ln \tilde q(\bx_n) = \Esub{q(\bx_{\backslash n})} {\ln p(\bx)} 
\end{empheq}
Implementing these update formulas does not require that the divergence that is being minimized ever needs to be evaluated. However, 
the fact that the divergence is guaranteed to decrease whenever one of the factors is updated is very useful for debugging.

In machine learning, the terms ``mean field approximation'' and ``variational inference'' are generally used interchangeably. As in the calculus of variations, the mean field updates (\ref{eqn:VBupdate})
are derived without imposing any conditions the functional forms of the factors in (\ref{eqn:17}) even 
in the case of continuous distributions. Hence the epithet ``variational".  The term ``inference'' generally refers to calculating exact or approximate 
posterior distributions of hidden variables in a probability model. (We will see how the mean field approximation accomplishes this in the next section.) However, as we have presented it, 
the mean field approximation can
be applied to the problem of calculating a factorized approximation $q(\bx)$ to
{\em any} target distribution $p(\bx)$ {\em provided that this problem is formulated as one of minimizing the reverse divergence $\KL{q(\bx)}{p(\bx)}$}. We will see in Section \ref{sec:P/A} how variational methods can be 
brought to bear on active inference as well as perceptual inference by 
casting the problem of inferring predictive distributions as one of optimizing a (suitably constrained) KL divergence rather than an expected free energy functional.

\subsection{Variational Free Energy}
\label{sec:VFE}
Suppose now that some of the components of $\bx$ are observable but others are not. In keeping with the notation that we will use later on, we set
$\bx = (\bs, \bo)$ where $\bo$ stands for the observable variables and $\bs$ for the hidden variables. 
Here we explain how variational inference can be used to derive an approximation to the posterior distribution $p(\bs|\bo)$ in situations where the exact posterior is intractable.

Suppose that $\bo$ is observed to take the value
$\bar \bo$. As we saw in Section \ref{sec:entropy}, the exact posterior $p(\bs|\bo)$ can be viewed as the solution to the problem of minimizing the divergence $\KL{q(\bs, \bo)}{p(\bs, \bo)}$
subject to the constraint that $q(\bs, \bo) = 0$ unless $\bo = \bar \bo$ or, equivalently, that $q(\bs, \bo)$ can be written in the form
$$q(\bs, \bo) = q(\bs) \delta_{\bar \bo} (\bo).$$
If $q(\bs)$ is required to factorize as in (\ref{eqn:17}), then the global minimum of the divergence may not be attainable
but the mean field algorithm returns an approximation to the exact posterior whose quality can be measured by the numerical value of the divergence. This numerical value
is known as the {\em variational free energy} (VFE):
\setlength\fboxsep{0.1cm}
\begin{empheq}[box=\widefbox]{align}
\label{eqn:VFE=KL}
{\rm VFE} = \KL{q(\bs, \bo)}{p(\bs, \bo)} \text{ where } q(\bs, \bo) = q(\bs)\delta_{\bar \bo}(\bo).
\end{empheq}

This is the most convenient way of defining variational free energy for our purposes but we can use the chain rule  (\ref{eqn:chain}) to express it in more familiar ways. 
The divergence 
$
\KL{q(\bs, \bo)}{p(\bs, \bo)}
$
can be written in either of the forms
\begin{align*}
& \KL{q(\bs)}{p(\bs)}   + \Esub{q(\bs)}{\KL{q(\bo|\bs)}{p(\bo|\bs)}}  
\intertext{or}
& \KL{q(\bo)}{p(\bo)}  + \Esub{q(\bo)}{\KL{q(\bs|\bo)}{p(\bs|\bo)}}
\end{align*}
which simplify to
\begin{align}
& \KL{q(\bs)}{p(\bs)} +\Esub{q(\bs)}{ - \ln p(\bar \bo|\bs)} \label{eqn:VFE1} 
\intertext{and}
&  - \ln p({\bar \bo}) + \KL{q(\bs)}{p(\bs|\bar \bo)} \label{eqn:VFE2}
\end{align}
under the assumption that $q(\bs, \bo) = q(\bs) \delta_{\bar \bo} (\bo)$.
In the active inference literature, the first term in (\ref{eqn:VFE1}) is known as the {\em complexity} as it can be interpreted as penalizing distributions $q(\bs)$ whose divergence from
the prior $p(\bs)$ is large. The second term is referred to as the {\em accuracy} since it can be interpreted as the average 
error incurred in reconstructing $\bar \bo$ by sampling from the distribution $q(\bs)$. 

Since the second term in (\ref{eqn:VFE2}) is non-negative, the variational free energy bounds the surprise $-\! \ln p(\bar \bo)$ from above. So choosing $q(\bs)$ 
to minimize the variational free energy can serve as a proxy for minimizing the surprise in situations where the marginal distribution $p(\bo)$ is intractable.

\section{Analogies with Statistical Mechanics}
\label{sec:StatMech}

Particularly in his early writings on the Free Energy Principle, Friston frequently conflates variational free energy and thermodynamic free energy, 
the energy in a system which is available to perform physical work such as moving a piston. 
This move has confused many of his readers and it is not needed to understand active inference, but the analogies 
with statistical mechanics are so fruitful that it is worth making an effort to understand how far they can be pushed. 
The ideas presented in this section are speculative and they will not be needed later so readers who are not interested 
in this topic are invited to skip it.

A physicist using statistical mechanical methods to study the thermodynamics of a neural population or of the brain as a whole
would aim to write down an energy function $E(\bx)$ which would, at least in principle, enable her to calculate the energy of the system
given its microstate $\bx$. This energy function would have the property that  the 
probability distribution $p(\bx)$ of microstates when the system is in thermal equilibrium with its environment is a Boltzmann
distribution. That is, 
\bE
\label{eqn:Boltzmann1}
p(\bx) = \frac{1}{Z(\beta)} e^{-\beta E(\bx)}
\eE
where the parameter $\beta$ is related to the temperature $T$ of the system by
\[
\beta = \frac{1}{k_B T}
\]
($k_B$ is Boltzmann's constant).
Taking $\beta = 1$, the free energy of the system is the total energy minus the entropy. 
Calculating the total energy by averaging over microstates and idetntifying the thermodynamic entropy with the Shannon entropy $\ent{p(\bx)}$, the free 
energy $\F{p(\bx)}$ is given by the expression
\[
\Esub{p(\bx)}{E(\bx)}  - \ent{p(\bx)}
\]
which reduces to $-\!\ln Z$.
If the neural population is maintained in a non-equilibrium steady state 
by sensory stimulation from the outside world and $q(\bx)$  is the distribution of microstates under this condition, then the non-equilibrium free energy $\F{q(\bx)}$ is given by
\[
\Esub{q(\bx)}{E(\bx)}  - \ent{q(\bx)}
\]
which can be written as 
$$ 
\KL{q(\bx)}{p(\bx)} - \ln Z.
$$
So the equilibrium and non-equilibrium free energies are related by
\[
\F{q(\bx)} = \KL{q(\bx)}{p(\bx)} + \F{p(\bx)}. 
\]
Thus the divergence $\KL{q(\bx)}{p(\bx)}$ can be interpreted as 
the additional thermodynamic free energy that is dissipated as heat
to the environment as the system relaxes towards equilibrium. Regarded in this way, the imperative to minimize the divergence from 
$q(\bx)$ to $p(\bx)$ is mandated by the second law of thermodynamics. 

We have assumed that the non-equilibrium steady state is the result of external simulation of sensory neurons so that some of the components of $\bx$ are fixed. 
We can restate this assumption in the notation introduced in Section \ref{sec:VFE}: $\bx = (\bs, \bo)$ and $\bo$ takes the value $\bar \bo$, so that  
\begin{align*}
\KL{q(\bx)}{p(\bx)} &= \KL{q(\bs, \bo)}{p(\bs, \bo)}
\intertext{and}
q(\bs, \bo) &= q(\bs) \delta_{\bar \bo} (\bo).
\end{align*}
So by (\ref{eqn:VFE=KL}), the divergence $\KL{q(\bx)}{p(\bx)}$ can be interpreted as a variational free energy as well as a thermodynamic free energy.

The Boltzmann distribution arises as the solution of a constrained optimization problem, namely the problem of finding the maximum entropy distribution on microstates for which the expected value of $E(\bx)$ coincides with a given observed value. 
(The parameter $\beta$ arises as a Lagrange multiplier.) More generally, given similar constraints on a collection of energy functions $\{ E^{(c)}(\bx) | c = 1, \ldots, C\}$
the maximum entropy distribution is a generalized Boltzmann distribution of the form
\begin{equation}
\label{eqn:Boltzmann}
p(\bx) = \frac{1}{Z(\bbeta)} \exp \left(- \sum_{c = 1}^C \beta^{(c)} E^{(c)}(\bx)\right)
\end{equation}
where $\bbeta = (\beta^{(1)}, \ldots, \beta^{(C)})$. This way of assigning probability distributions is known as the maximum entropy principle \cite{Jaynes1957InformationTheoryStatMech,Jaynes1957InformationTheoryStatMech2}. Since maximizing the entropy of a distribution is equivalent to minimizing its 
divergence from a uniform distribution, the maximum entropy principle can be regarded as another instance of constrained divergence minimization.

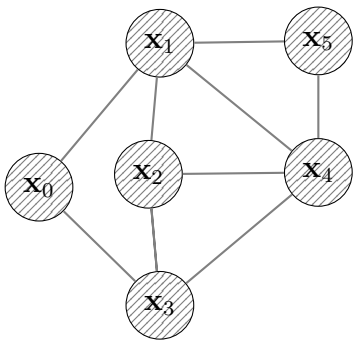
\begin{figure}[htb]
\centering
\begin{tikzpicture}[bn/.style={circle,draw,fill=white, pattern=north east lines, pattern color=gray=white,text=black,font=\sffamily,minimum
size=9mm},every node/.append style={bn}]
 \path node (1) {${\bx_0}$} -- ++ (50:2.5) node (2) {${\bx_1}$} -- ++(-95:1.75) node (3) {${\bx_2}$}
 -- ++(-85:1.75) node (4) {${\bx_3}$} -- ++(40:2.75) node (5) {${\bx_4}$}
 -- ++ (0,1.75) node (6) {${\bx_5}$} ;
 \draw[gray, thick] (1)--(2)--(6)--(5)--(4)--(3)--(5)--(2)--(3)--(4);
 \draw[gray, thick] (1)--(4);
\end{tikzpicture}
\caption{\em A toy model of the brain as a Markov Random Field. Nodes correspond to grey matter and branches to white matter. Energy functions are associated with both nodes and branches. Energy functions associated with the nodes model the microstates of neural populations.
Energy functions associated with the branches model the dependencies between the microstates of the neural populations associated with the nodes. One or more of the nodes receives sensory data from the external world which changes from moment to moment. The neural populations are continually striving to dissipate free energy by exciting or inhibiting the populations with which they are in communication via the branches.}
\label{fig:MRF}
\end{figure}

Considering that all of the probability distributions in statistical mechanics have the form of generalized Boltzmann distributions, the question arises: Can the brain as a whole be modelled in this way? For this we would need to define a collection of energy functions which are localized in the sense that
each of them of is a function of the microstates of a (relatively small) neuronal population. For the sake of argument, let us imagine representing the brain by a sparsely connected graph as in Fig. \ref{fig:MRF}.  
The nodes represent neural populations (grey matter) and the branches represent communication channels between populations (white matter). The variables $\bx_0, \bx_1, \ldots$ associated with the nodes are the microstates of the corresponding populations. 
We could associate energy functions  $E^{(0)}(\bx_0), E^{(1)}(\bx_1), \dots$ with each of the nodes and define a probability distribution by invoking (\ref{eqn:Boltzmann}) but this would result in a distribution under 
which the variables $\bx_0, \bx_1, \ldots$ are statistically independent which is obviously inadequate. In the case of a Gaussian Markov random field, this could be corrected by adding off-diagonal terms to the precision matrix. 
Analogously, 
we could model the dependencies between neural populations by adding energy functions depending on two sets of variables for each branch in the graph. 
For example, the branch joining node 0 and node 3 would contribute an energy function $E^{(03)}(\bx_0, \bx_3)$ whose role is to model the dependency between $\bx_0$ and $\bx_3$.  

We assume that one of nodes in the graph is distinguished from the others by being in communication with the world outside the brain (which includes the body as well as the environment). 
The neural population corresponding to this node consists of sensory neurons and the microstate of this population at a given time depends on what is going on in the world as well as the microstates of the populations 
with it which communicates. In this picture, the ``brain'' tracks changes in the body and environment by continually updating approximate variational posterior distributions for each of the neural populations.

An attractive feature of this model is that it suggests that a similar picture obtains at finer scales of resolution. Implementing the mean field update formula \ref{eqn:VBupdate} for each of the 
nodes in the graph Fig. \ref{fig:MRF} requires reading off the variational posteriors associated with the nodes with which it communicates and only those nodes. 
For example, the variational posterior $q(\bx_0)$ is updated by calculating the energy function $\Esub{q(\bx_{\backslash 0})}{E(\bx)}$ which is given (up to an irrelevant additive constant) by
\[
E^{(0)}(\bx_0) + \Esub{q(\bx_1)}{E^{(01)}(\bx_0, \bx_1)} +  \Esub{q(\bx_3)}{E^{(03)}(\bx_0, \bx_3)}. 
\]
So nodes 1 and 3 stand in the same relationship to node 0 as the sensory neurons to the ``brain'' as a whole. Thus if the neural population corresponding to node 0 admits of a graphical representation similar to Fig. \ref{fig:MRF}, then the mechanics of updating the local variational posterior $q(\bx_0)$ will be similar to that of updating the global variational posterior $q(\bx_0, \bx_1, \ldots)$. The same sort of analysis may be applicable to neural subpopulations at different scales, perhaps even down to the scale of individual neurons.

There are no restrictions on the topology of the graph in Fig. \ref{fig:MRF}. Variational free energy is continually being minimized by passing messages between communicating neural populations 
but this does not require that the variational free energy of the brain as a whole be evaluated from moment to moment. Hence there is no need
for a ``central processing unit" or neo-Cartesian ``inner screen"  \cite{ramstead_inner_2024} in the picture we have sketched. All that is going on is that each neural population and 
subpopulation is continually striving to dissipate its own free energy by exciting and/or inhibiting the activity of the other populations with which it communicates.

The Lagrange multipliers in (\ref{eqn:Boltzmann1}) and (\ref{eqn:Boltzmann}) have an interesting role to play. Decreasing $\beta$  (that is to say, increasing the temperature) in (\ref{eqn:Boltzmann1}) has the effect of increasing the entropy of the distribution $p(\bx)$ which quantifies 
the uncertainty about the value that $\bx$ actually takes. This operation is usually referred to as {\it precision weighting} since the effect of applying it to a Gaussian distribution is to leave the mean unchanged but scale the precision (or inverse variance) by a factor of $\beta$.  Similarly, changing the
multipliers associated with the branches in the graph in Fig. \ref{fig:MRF} has the effect of modulating the strength of the interactions between neural populations.  
Precision weighting is thought to be the mechanism underlying attention \cite{Feldman2010} and arousal modulation \cite{solms2021_hiddenspring} and the computational neuroscience literature abounds in (more or less plausible) 
just-so stories which use this mechanism to explain neurological anomalies such as aphantasia, akinesia and apathy in dementia, and the sensory effects of psychedelics.

An agent cannot act effectively on its environment if it is uncertain about the current state of its world or about the accuracy of its predictions of future states. 
In such a situation actions have to be guided by the imperative to reduce uncertainty. (We will discuss the exploration/exploitation tradeoff in Section \ref{sec:info}.) 
So an intelligent agent needs to assess the degree of confidence it has in its inferences and predictions before it acts in a given situation. Treating the Lagrange multipliers as random variables subject to probabilistic inference would endow 
the agent with the flexibility needed to do this well. This topic has hardly begun to be explored in the active inference literature (\cite{Sandved-Smith2021Towards} is an exception 
but the ``universal generative model'' of \cite{friston_supervised_2023, friston_pixels_2024} does not include precision weighting).  
So we will not dwell on it here except  to point out that there doesn't seem to be any major obstacles in the way of a fully probabilistic treatment should that prove to be useful. 

For example, if we assume a joint distribution on microstates and multipliers of the form
\[
p(\bbeta, \bx) \propto \exp \left(- \sum_{c = 1}^C \beta^{(c)} E^{(c)}(\bx)\right)
\]
and a variational approximation of the form
\[
q(\bbeta, \bx) = \prod_c q(\beta^{(c)}) \prod_n q(\bx_n)
\]
the factors $q(\beta^{(c)})$ turn out to be exponential distributions. The assumption here that the $\beta^{(c)}\/$'s are statistically independent in the approximating distribution is arguably unsatisfactory
if the intention is to model attention since attention is usually thought of as a limited resource which is concentrated on one object at a time. This objection might be answered by assuming a joint distribution of the form 
\[
p(\bmu, \bx) \propto \prod_c \left( \mu^{(c)} \right)^{E^{(c)}(\bf x)}
\]
where the $\mu^{(c)}\/$'s are non-negative and sum to 1. Under these conditions, a variational approximation of the form
\[
q(\bmu, \bx) = q(\bmu) \prod_n q(\bx_n)
\]
yields a Dirichlet distribution for $q(\bmu)$. (The Dirichlet distribution is discussed in Section \ref{sec:Dirichlet}.)

\section{Directed Graphical Models}
\label{sec:networks}

Energy based models lend themselves naturally to variational approximation 
methods but they are difficult to train in 
situations where the partition function cannot be evaluated and this is generally 
the case if the state space is of high dimension. For this reason, probability distributions
are not usually specified by energy functions in machine learning but by directed graphical models, also known as Bayesian networks. 

\subsection{A Toy Model of Predictive Processing in the Visual Cortex }
\label{sec:toy}

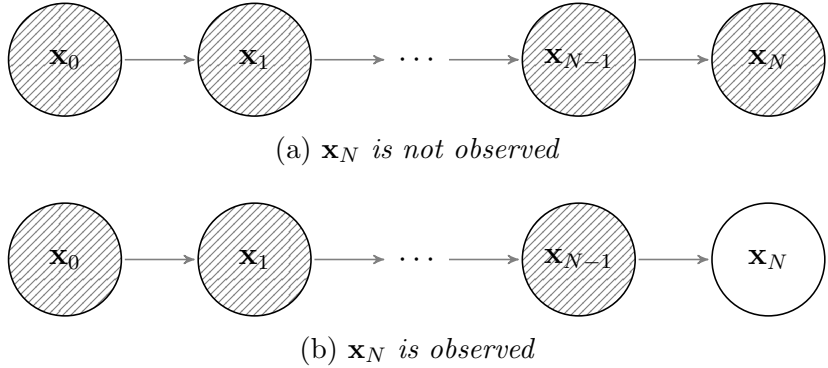
\begin{figure}[htbp]
        \centering
        \begin{subfigure}[b]{0.4\textwidth} 
        \centerline{
	\begin{tikzpicture}[modal]
		\node[world] (1) {$\bx_0$};
		\node[world] (2) [right=of 1] {$\bx_1$};	
		\node []             (3) [right=of 2] {$\mathbf \cdots$};
		\node[world] (4) [right=of 3] {$\bx_{N-1}$};	
		\node[world] (5) [right=of 4] {$\bx_{N}$};	
		
		\path[->,gray] (1) edge (2);
		\path[->,gray] (2) edge (3);
		\path[->,gray] (3) edge (4);
		\path[->,gray] (4) edge (5);		
	\end{tikzpicture}
	}
\caption{\em $\bx_N$ is not observed}
    \label{subfig:image1}
    \end{subfigure}
    \par\vspace{1em}
    
\begin{subfigure}[b]{0.4\textwidth} 
        \centerline{
	\begin{tikzpicture}[modal]
		\node[world] (1) {$\bx_0$};
		\node[world] (2) [right=of 1] {$\bx_1$};	
		\node []             (3) [right=of 2] {$\mathbf \cdots$};
		\node[world] (4) [right=of 3] {$\bx_{N-1}$};	
		\node[point] (5) [right=of 4] {{$\bx_{N}$}};	
		
		\path[->,gray] (1) edge (2);
		\path[->,gray] (2) edge (3);
		\path[->,gray] (3) edge (4);
		\path[->,gray] (4) edge (5);		
	\end{tikzpicture}
	}
\caption{\em $\bx_N$ is observed}
    \end{subfigure}
    \par\vspace{1em}   
        \caption{\em A toy model for predictive processing in the visual cortex. Hatching indicates variables whose values are not observed.}
        \label{fig:PP}
    \end{figure}

For example, the graph in Fig. \ref{fig:PP} represents a toy model of hierarchical predictive 
processing in the visual cortex \cite{Millidge2021PredictiveCA}. The variable $\bx_N$ represents the state of the retina, the variable
$\bx_{N-1}$ represents the proximate causes of $\bx_N$ and the directed arrow joining the two indicates that the causal relationship is specified
by a conditional probability distribution of the form $p(\bx_N|\bx_{N-1})$, and so on up the hierarchy. 
So if we set $\bx = (\bx_0, \ldots, \bx_N)$, the assumption is that  the joint distribution $p(\bx)$ 
factorizes as a product of conditional probability distributions
\bE
p(\bx) = p(\bx_0) p(\bx_1|\bx_0) \cdots p(\bx_N|\bx_{N-1}).
\eE
As a warm-up exercise for the calculations that we will perform in Section \ref{sec:P/A} and the appendix, 
consider the problem of approximating $p(\bx)$ by a factorized
distribution of the form
\bE
\label{eqn:toy}
q(\bx) = q(\bx_0) q(\bx_1) \cdots q(\bx_{N}) 
\eE
by variational methods. In the case where $\bx_N$ is not observed (panel (a) of Fig. \ref{fig:PP}), we update all of the factors one at time so that, on convergence, $q(\bx_N)$ 
is an approximation to the marginal distribution $p(\bx_N)$.
In the case where $\bx_N$ is observed (panel (b) of Fig. \ref{fig:PP}) we impose the additional constraint that the distribution $q(\bx_N)$ is concentrated on the observed value $\bar \bx_N$.
So $q(\bx_N)$ is never updated and, on convergence,  $q(\bx)$ is an approximation to the posterior distribution $p(\bx|\bx_N = \bar \bx_N)$.

If we assume that for $n = 0, \ldots, N$, $\bx_n$ can take $K$ possible values, 
then we can represent $\bx_n$ as a 1-hot $K \times 1$ vector and write
\begin{align*}
p(\bx_0) &= \bx^\top_0 \bp_0 \\
q(\bx_n) &= \bx^\top_n \bq_n \\
p(\bx_n | \bx_{n-1}) &= \bx^\top_{n - 1} \bA_n \  \bx_n
\end{align*}
where $\bp_0$ and $\bq_n$ are $K \times 1$ vectors and $\bA_n$ is a $K \times K$ matrix. Thus
\[
\ln p(\bx) =  \bx^\top_0 \ln \bp_0 + \sum_{n = 1}^N  \bx^\top_{n - 1} \ln \bA_n \  \bx_n
\]
where the logarithms of the vectors and matrices on the right-hand side are taken element wise.

Recall that by (\ref{eqn:VBupdate}), the recipe for updating one of the factors $q(\bx_n)$ in (\ref{eqn:toy}) holding the others fixed
calls for evaluating $\tilde q(\bx_n)$ defined by
$$
\ln \tilde q(\bx_n) = \Esub{q(\bx_{\backslash n})}{\ln p(\bx)}.
$$
This gives
\begin{align*}
\ln \tilde q(\bx_n) = \begin{cases}
\bx^\top_0 \left(\ln \bp_0 + \ln \bA_{1}\ \bq_1 \right) + \ldots & (n = 0) \\
\bx^\top_n \left(  \ln \bA^\top_n \bq_{n -1}  +  \ln  \bA_{n + 1}  \ \bq_{n+1} \right) + \ldots  & (0 < n < N)\\
 \bx^\top_N \ln \bA^\top_{N-1}  \bq_{N}  + \ldots & (n = N)
\end{cases}
\end{align*}
where the ellipses indicate terms which are independent of $\bx_n$ and hence can be ignored.\footnote{We use the convention $+ \ldots$ to indicate irrelevant additive terms throughout.}
(We have used the fact that the scalar 
$\bq^\top_{n-1} \ln \bA_n \ \bx_{n}$ can be rewritten as 
$\bx^\top_n \ln \bA^\top_n \bq_{n -1}$.)
So the update formula for $\bq_n$ is
\begin{align*}
\ln \bq_n = 
\begin{cases}
\ln \bp_0 + \ln \bA_{1} \bq_1 + \ldots & (n = 0) \\
 \ln \bA^\top_n \bq_{n -1}  +  \ln  \bA_{n + 1}   \bq_{n+1}  + \ldots  & (0 < n < N)\\
 \ln \bA^\top_{N-1}  \bq_{N}  + \ldots & (n = N)
\end{cases}
\end{align*}
where the ellipses indicate constants whose role is to ensure that the components of $\bq_n$ sum to 1. In other words, for each $n = 0, \ldots, N$, 
the vector $\bq_n$ is derived from the corresponding vector on the right-hand side by passing it through the softmax function. 
In the case were $\bx_N$ is observed, the update formula is not applied in the case $n = N$, and $\bq_N$ is set to be the 1-hot vector corresponding to 
the observed value of $\bx_N$.

To evaluate the divergence $\KL{q(\bx)}{p(\bx)}$, note that
\begin{align*}
\Esub{q(\bx)} {\ln p(\bx)} &=  \bq^\top_0 \ln \bp_0 + \sum_{n = 1}^N  \bq^\top_{n - 1} \ln \bA_n \ \bq_n \nonumber \\
\intertext{and} \Esub{q(\bx)} {\ln q(\bx)} &= \sum_{n=0}^N \bq_n \ln \bq_n \\
\intertext{so that}
\KL{q(\bx)}{p(\bx)} 
&= \sum_{n = 0}^N \bq^\top_n \ln \bq_n -   \bq^\top_0 \ln \bp_0 -  \sum_{n = 1}^N  \bq^\top_{n - 1} \ln \bA_n \ \bq_n.
\end{align*}



%
\subsection{Static and Dynamic Bayesian Networks}
\begin{figure}[h]
\centering
\scalebox{0.7}{
\begin{tikzpicture}[
  level distance=3cm,
  every node/.style={circle, draw, minimum size=10mm, inner sep=0pt},
  level 1/.style={sibling distance=5cm},
  level 2/.style={sibling distance=2.5cm},
  level 3/.style={sibling distance=1.25cm},
  edge from parent/.style={draw,gray,->,thick}
]

\node[fill=white, pattern=north east lines, pattern color=gray] {}
  child { node[fill=white, pattern=north east lines, pattern color=gray] {}
    child { node[fill=white, pattern=north east lines, pattern color=gray] {}
      child { node[] {} }
      child { node[] {} }
    }
    child { node[fill=white, pattern=north east lines, pattern color=gray] {}
      child { node[] {} }
      child { node[] {} }
    }
  }
  child { node[fill=white, pattern=north east lines, pattern color=gray] {}
    child { node[fill=white, pattern=north east lines, pattern color=gray] {}
      child { node[] {} }
    }
    child { node[fill=white, pattern=north east lines, pattern color=gray] {}
      child { node[] {} }
      child { node[] {} }
      child { node[] {} }
    }
  }
  child { node[fill=white, pattern=north east lines, pattern color=gray] {}
    child { node[fill=white, pattern=north east lines, pattern color=gray] {}
      child { node[] {} }
    }
    child { node[fill=white, pattern=north east lines, pattern color=gray] {}
      child { node[] {} }
      child { node[] {} }
    }
  };

\end{tikzpicture}
}
\caption{\em A toy model for simultaneous predictive processing in multiple sensory modalities. The leaf nodes of the tree correspond to sensors.}
\label{fig:tree}
\end{figure}
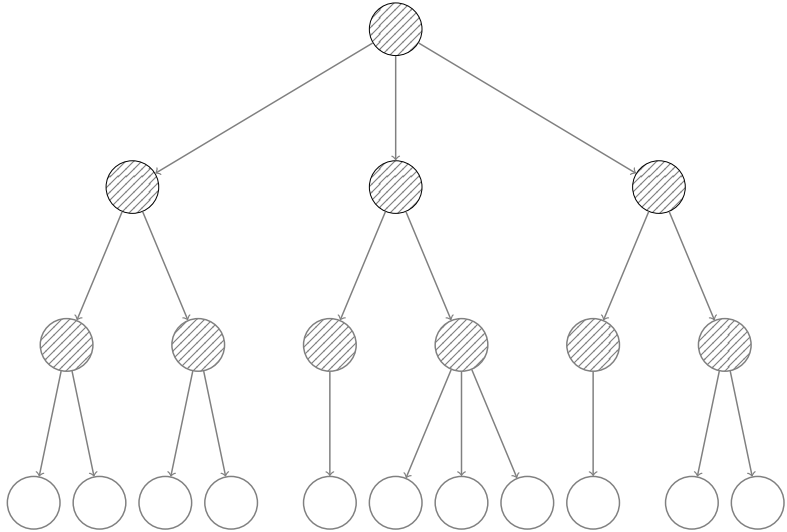

A tree structure as in Fig. \ref{fig:tree}, rather than the linear structure in Fig. \ref{fig:PP}, enables our toy predictive processing model to be extended to embrace multiple sensory modalities. Associated with each node $n$ in the tree there is a variable $\bx_n$, and a conditional probability
distribution $p(\bx_{n'}|\bx_n)$ is specified for each branch $n \rightarrow n'$. Think of the leaf nodes of the tree as the various sense gates and of the variables associated with nodes at higher levels as progressively more abstract representations of the sensorium. 

Variational message passing in a tree works in much the same 
way as in the linear graph Fig. \ref{fig:tree}: updating the variational posterior associated with a given node depends on messages received from its parent and its children (rather than from its successor and predecessor as in the case of Fig. \ref{fig:PP}) but not from any other nodes.

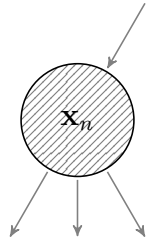
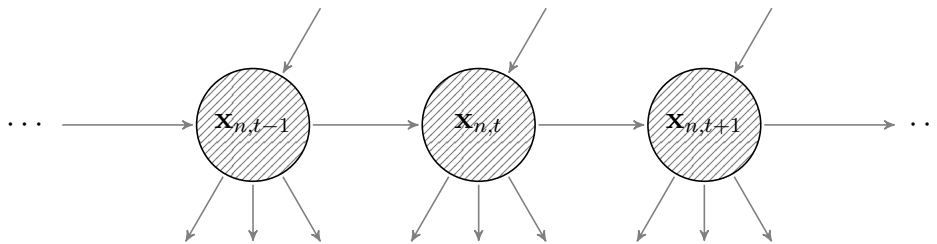
\begin{figure}[h]
 \centering
    \begin{tabular}{c}
    \begin{subfigure}[c]{0.45\textwidth}
        \centerline{
	\begin{tikzpicture}[modal]
		\node[world]  (a1) {$\bx_n$};
		\node[]  (a2) at ($(a1) + (60:2)$) {};
		\node[]  (a3) at ($(a1) + (240:2)$) {};
		\node[]  (a4) at ($(a1) + (270:1.73)$) {};
		\node[]  (a5) at ($(a1) + (300:2)$) {};
		
		\path[->,gray]  (a2) edge  (a1);
		\path[->,gray]  (a1) edge  (a3);
		\path[->,gray]  (a1) edge  (a4);
		\path[->,gray]  (a1) edge  (a5);	
	\end{tikzpicture} 
	}              
        \caption{\em Static Network.}
        \label{subfig:one}
    \end{subfigure}
    
    \hfill \\
    
    \centering
    
    \begin{subfigure}[c]{0.45\textwidth}
        \centerline{
        	\begin{tikzpicture}[modal]
		\node[world]  (a1) {$\bx_{n, t-1}$};
		\node[world]  (b1)  at ($(a1) + (0:3)$)  {$\bx_{n, t}$};
		\node[world]  (c1)  at ($(b1) + (0:3)$) {$\bx_{n, t+1}$};
		
		\node[]  (x)  at ($(a1) + (180:3)$)  {$\cdots$};
		\node[]  (y)  at ($(c1) + (0:3)$) {$\cdots$};
			
		\path[->,gray] (x) edge (a1);
		\path[->,gray] (a1) edge (b1);
		\path[->,gray] (b1) edge (c1);
		\path[->,gray] (c1) edge (y);

		\node[]  (a2) at ($(a1) + (60:2)$) {};
		\node[]  (a3) at ($(a1) + (240:2)$) {};
		\node[]  (a4) at ($(a1) + (270:1.73)$) {};
		\node[]  (a5) at ($(a1) + (300:2)$) {};
		
		\path[->,gray]  (a2) edge  (a1);
		\path[->,gray]  (a1) edge  (a3);
		\path[->,gray]  (a1) edge  (a4);
		\path[->,gray]  (a1) edge  (a5);	
		
		\node[]  (b2) at ($(b1) + (60:2)$) {};
		\node[]  (b3) at ($(b1) + (240:2)$) {};
		\node[]  (b4) at ($(b1) + (270:1.73)$) {};
		\node[]  (b5) at ($(b1) + (300:2)$) {};
		
		\path[->,gray]  (b2) edge  (b1);
		\path[->,gray]  (b1) edge  (b3);
		\path[->,gray]  (b1) edge  (b4);
		\path[->,gray]  (b1) edge  (b5);	
		
		\node[]  (c2) at ($(c1) + (60:2)$) {};
		\node[]  (c3) at ($(c1) + (240:2)$) {};
		\node[]  (c4) at ($(c1) + (270:1.73)$) {};
		\node[]  (c5) at ($(c1) + (300:2)$) {};
		
		\path[->,gray]  (c2) edge  (c1);
		\path[->,gray]  (c1) edge  (c3);
		\path[->,gray]  (c1) edge  (c4);
		\path[->,gray]  (c1) edge  (c5);	
	
	\end{tikzpicture}  
	}
        \caption{\em Dynamic Network.}
        \label{subfig:two}
    \end{subfigure}
     \end{tabular}

\caption{\em Constructing a Dynamic Bayesian Network from a static network. The static network supports prediction from top to bottom (as in figs. \ref{fig:PP} and \ref{fig:tree}). The dynamic network supports prediction from past to future as well.}
\label{fig:DBN}
\end{figure}

Oddly enough, the term ``predictive processing" in the neuroscience literature is primarily used to refer to prediction from top to bottom in a hierarchical network such as Figs. \ref{fig:PP} and \ref{fig:tree} rather than prediction from past to future which 
is evidently just as important. Fig. \ref{fig:DBN} illustrates a general procedure for converting a given Bayesian Network to a {\em Dynamic} Bayesian Network which enables temporal as well as top down prediction. 

The construction consists in replacing a node in the original network representing a random variable $\bx_n$ 
by a chain of nodes representing instances of $\bx_n$ at successive times. This entails replacing the conditional distribution $p(\bx_{n'}|\bx_n)$ associated with a given branch in the original graph
by a conditional distribution of the form  $p(\bx_{n', t}|\bx_{n, t}, \bx_{n', t - 1})$ for each time $t$. This  probability distribution  
is usually taken to be independent of $t$, although we will encounter situations where time-dependent transition probabilities are needed.


An important point to note is that this construction can be implemented with slower clock speeds at higher levels of the hierarchy so that the various 
levels can be thought of as representing the sensorium on different temporal scales as well as on different spatial scales.  Philosophy of mind has traditionally been stumped by the binding problem in sense perception but this schema shows how  
the Bayesian Brain Hypothesis deals with it straightforwardly.

%
%
%
%
%

\subsection{Hidden Markov Models}

Henceforth, we will use the notation $\bs$ to indicate a {\em sequence} of hidden variables or ``states" $\bs_0, \bs_1, \ldots, \bs_T$ in a Dynamic Bayesian Network, $\bo$ to indicate a {sequence} of observable variables $\bo_1, \ldots, \bo_T$
and $\bar \bo_1, \ldots, \bar \bo_T$ to indicate a sequence of observed values. For each time $t = 1, \ldots, T - 1$, $\bs_{> t}$ denotes the sequence $\bs_{t + 1}, \ldots, \bs_T$ and so forth.

A {\em Hidden Markov Model} (HMM) is a joint probability distribution $p(\bs, \bo)$ on state sequences $\bs$ and observation sequences $\bo$ of the form
\bE
p(\bs, \bo) 
= p(\bs_0) \prod_{t = 1}^T p(\bs_t|\bs_{t - 1}) p(\bo_t|\bs_t). \label{eqn:HMM}
\eE
So the sequence of states is a Markov chain and, at each time $t$, $\bo_t$ is predicted from $\bs_{t}$ alone rather than from the pair $(\bs_{t}, \bo_{t -1})$ 
(as would be the case in more  general Dynamic Bayesian Networks). The intuition here is that sensory data (as distinct from, say, the text data which large language models have to predict) is so noisy that
 useful predictions are only possible in the space of hidden variables (or embeddings, to use the language of artificial intelligence \cite{Dawid_2024,hafner_mastering_2024}). 
 In a good probability model, the state space will be of lower dimension than the observation space so that inferring states from observations can be thought of as a denoising operation which 
 can be expected to enhance the accuracy of perceptions and predictions.

\section{The Perception/Action Cycle}
\label{sec:P/A}

Recall the basic tenets of the Bayesian Brain Hypothesis: A biological agent infers the state of its limbs and viscera at a given time from
its proprioceptive and interoceptive sensations and the state of its environment from its exteroceptive sensations. This is known as perceptual inference. It consists in calculating an approximate posterior distribution of the hidden variables in the agent's model of its world by minimizing variational free energy.  The agent navigates its world by calculating a predictive distribution on its future sensations. This is known as active inference. If active inference is to be cast as an optimization problem in the same way as perceptual inference then an objective function analogous to variational free energy needs to be defined. For an agent that models its world as a HMM, we propose that 
the divergence of a predictive distribution from the HMM can serve as such an objective function.  

We have already seen in Sections \ref{sec:VFE} and \ref{sec:toy} how calculating 
approximate posterior and predictive distributions can be formulated as constrained divergence minimization problems that can be solved by variational methods. In this section we show how a suitably constrained mean field 
approximation to the HMM that models the agent's world can serve both as a posterior distribution for perceptual inference and as a predictive distribution for active inference. 

When it is used to model perception, the divergence criterion that we optimize coincides with variational free energy. When it used to model action, it turns out to be closely related to but different from the expected free energy functionals that have previously been used in active inference.
We will discuss this relationship in  Section \ref{sec:whither} below. 

\subsection{Perception and Action as Divergence Minimization}
\label{sec:P/A.1}

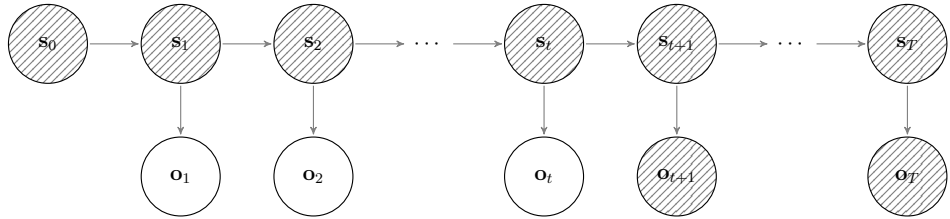
\begin{figure}[h]
\centering
\scalebox{0.7}{
	\begin{tikzpicture}[modal]
		\node[world](s0){$\bs_0$};
		\node[world] (s1)[right=of s0] {$\bs_1$};
		\node[world] (s2) [right=of s1] {$\bs_2$};	
		\node []         (s3) [right=of s2] {$\cdots$};		
		\node[world] (s4) [right=of s3]  {$\bs_{t }$};
		\node[world] (s5) [right=of s4]  {$\bs_{t + 1}$};
		\node []         (s6) [right=of s5] {$\cdots$};		
		\node[world] (s7) [right=of s6]  {$\bs_{T}$};
		
		\path[->,gray](s0) edge (s1);
		\path[->,gray](s1) edge (s2);
		\path[->,gray](s2) edge (s3);
		\path[->,gray](s3) edge (s4);
		\path[->,gray](s4) edge (s5);		
		\path[->,gray](s5) edge (s6);
		\path[->,gray](s6) edge (s7);

		\node[point] (o1) [below=of s1] {{$\bo_1$}};
		\node[point] (o2) [below=of s2] {{$\bo_2$}};
		\node[point] (o3) [below=of s4] {{$\bo_{t}$}};
		\node[world] (o4) [below=of s5] {$\bo_{t + 1}$};
		\node[world] (o5) [below=of s7] {$\bo_T$};
		
		\path[->,gray](s1) edge (o1);
		\path[->,gray](s2) edge (o2);
		\path[->,gray](s4) edge (o3);
		\path[->,gray](s5) edge (o4);
		\path[->,gray](s7) edge (o5);
		
	\end{tikzpicture}
}
\caption{Directed graphical model for the Perception/Action cycle. At time $t$, $\bo_{\le t}$ has been observed and  $\bo_{> t}$ is yet to be observed. }
\label{fig:PA}
\end{figure}

The title of this section is borrowed from \cite{hafner_action_2022} where it is shown how a wide range of reinforcement learning algorithms can be viewed as 
solving constrained divergence minimization problems. Given a HMM  $p(\bs, \bo)$ and an observation history $\bar \bo_1, \ldots, \bar \bo_{t}$,
we show how perceptual and active inference can be cast as a problem of optimizing a joint divergence 
functional $\KL{q(\bs, \bo)}{p(\bs, \bo)}$ where
$q(\bs, \bo)$ a suitably constrained variational approximation to ${p(\bs, \bo)}$. Our approach enables us to treat the perception/action cycle in a unified way
rather than modelling perception by optimizing a variational free energy functional and action by optimizing an expected free
energy functional (as is usually done).

The graphical model for the calculation we will perform is shown in fig. \ref{fig:PA}. 
We assume that the planning horizon $T$ is fixed in advance. At a given time $t = 0, \ldots, T$, the variables $\bo_1, \ldots, \bo_{t}$ 
have been observed and we use the observed values  $\bar \bo_1, \ldots, \bar \bo_{t}$ to 
calculate posterior distributions for the entire state sequence $\bs_0, \ldots, \bs_T$ up to time $T$. So, at each time $t$, we are retrodicting the past as well as predicting the future.
Retrodiction has to be supported if an agent is to have the ability to update its beliefs about past events as more information becomes available.

Following the standard convention in the active inference literature,  we use $t$ to denote the {\em present} time and $\tau$ to denote any time (past, present, or future) in the interval $0, \ldots, T$. For a given present time $t$, 
we assume that $q(\bs, \bo)$ factorizes as 
\begin{align*}
q(\bs, \bo)  &= q(\bs_0) \prod_{\tau = 1}^T q(\bs_\tau, \bo_\tau) 
\end{align*}
where $q(\bs_0) = p(\bs_0)$ and
\begin{align}
q(\bs_\tau, \bo_\tau) &= \begin{cases}
q(\bs_\tau) \delta_{\bar \bo_\tau}(\bo_\tau) &\quad (\tau = 1, \ldots, t)\\
q(\bs_\tau) p(\bo_\tau|\bs_\tau) &\quad (\tau = t + 1, \ldots T). 
\end{cases} \label{eqn:factorization}
\end{align}
Another (arguably better) possibility which we will not explore here would be to factorize $q(\bs)$ as a product of conditional distributions so that
\[
q(\bs) = q(\bs_0) \prod_{\tau = 1}^T q(\bs_\tau|\bs_{\tau-1}).
\] 
This is the most commonly used factorization in Bayesian treatments of HMMs \cite{Beal2003}. In the context of active inference, this factorization
is discussed in \cite{ParrMarkovicFriston2019,Parr2022ActiveInference} where it is referred to as the Bethe approximation. 

Note that because of the presence of the term $p(\bo_\tau|\bs_\tau)$ in (\ref{eqn:factorization}) 
the factorization we are using is not a full mean field approximation to $p(\bs, \bo)$. We will explain the reason for this choice in Section \ref{sec:info} below.

Note also that, although our notation fails to reflect these dependencies, the value of the divergence from $q(\bs, \bo)$ to $p(\bs, \bo)$ depends
on the starting distribution $p(\bs_0)$ (through (\ref{eqn:HMM})) as well as on the observations up to time $t$ (through (\ref{eqn:factorization})).  We will treat the starting 
distribution as context-dependent in the sense that it is determined 
by the agent's activity prior to time $0$ and we will take the variational posterior $q(\bs_0)$ to be identical to $p(\bs_0)$. Equation (\ref{eqn:G2}) below explains the rationale for this choice. 
Our handling of the starting distribution departs from the tradition in the active inference literature where it is specified by a vector denoted by $\bD$ which is assumed to be context-independent
and the variational posterior at time 0 is updated in the same way as the variational posteriors at other times.


Throughout this section, we ignore any structure that the state space of the HMM may have and we assume that the number of states sufficiently small that the emission probability distributions and 
transition probability distributions that define the HMM can be precomputed and stored as matrices which we denote by $\bA$ and $\bB$ respectively (following the conventions in the active inference literature). 

As in Section \ref{sec:toy} we represent $\bs_\tau$ and $\bo_\tau$ by 1-hot vectors so that
\begin{align}
p(\bo_\tau|\bs_\tau) &= \bs^\top_\tau \bA \bo_\tau \nonumber \\
p({\bs_\tau}|\bs_{\tau-1}) &= \bs^\top_{\tau-1} \bB \bs_\tau \nonumber \\
p(\bs_0) &= \bs^\top_0 \bp_0 \label{eqn:aa}
\end{align}
where $\bp_0$ is the vector that represents the starting distribution $p(\bs_0)$. With these conventions,
\bE
\label{eqn:joint}
\ln p(\bs, \bo) =  {\bs}^\top_0 \ln {\bp}_0  + \sum_{\tau = 1}^T \left( {\bs}_\tau^\top \ln \bA  \ {\bo}_\tau  +   {\bs}^\top_{\tau - 1} \ln \bB\ {\bs}_{\tau} \right) .
\eE
Likewise, we store each posterior distribution $q(\bs_\tau)$ as a vector $\bq_\tau$ so that 
\begin{align}
q({\bs_\tau}) &= \bs^\top_{\tau} \bq_\tau. \label{eqn:bb}
\end{align}

We derive the formulas for updating each posterior $\bq_\tau$ holding the others fixed in Sections \ref{predicting} and \ref{retrodicting} in the appendix. 
In the case where $\tau > t$, the update for $\bq_\tau$ is given by (\ref{eqn_app:update2}):
\begin{align}
\ln \bq_\tau &= 
\begin{cases}
\ln \bB^\top \bq_{\tau-1} + \ln \bB \ \bq_{\tau + 1}  + \ldots& \quad  (\tau  = 1, \ldots, T - 1) \\
\ln \bB^\top \bq_{T-1}  + \ldots & \quad (\tau = T).
\end{cases}\label{eqn:update2}
\end{align}
We refer to this as the {\em prediction} update formula.
In the case where $\tau \le t$, the update for $\bq_\tau$ is given by (\ref{eqn_app:update1}):
\begin{align}
\ln \bq_\tau
&= \begin{cases}
\ln \bp_0 & \quad (\tau = 0) \\
\ln \bA \  \bar \bo_\tau   + \ln \bB^\top \bq_{\tau-1} + \ln \bB \ \bq_{\tau + 1}  \ldots& \quad  (\tau  = 1, \ldots, T - 1) \\
\ln \bA \  \bar \bo_T + \ln \bB^\top \bq_{T-1}  +  \ldots & \quad (\tau = T).
\end{cases} \label{eqn:update1}
\end{align}
We refer to this as the {\em retrodiction} update formula. But for the fact that we have simplified the way the case $\tau = 0$ is handled, the retrodiction updates
are the standard updates for minimizing the variational free energy of the observations $\bar \bo_1, \ldots, \bar \bo_t$.  

Broadly speaking, there are two ways of implementing these update formulas at each present time $t$. Both implementations can be carried out in real time, 
but they have different computational requirements. 
The simplest, known as {Bayesian filtering} (by analogy with Kalman filtering), 
holds $\bq_0, \ldots, \bq_{t -1}$ fixed and updates the posteriors for the present and future times $\bq_{t}, \ldots, \bq_T$ asynchronously
at time $t$. The alternative, known as {Bayesian smoothing}, updates all of the posteriors $\bq_1, \ldots, \bq_T$ asynchronously at every time present time $t$. 
This results in a lower value for the divergence criterion. An obvious compromise between the two approaches would be to limit the extent to which beliefs about the past
are updated in Bayesian smoothing.

After calculating the posteriors at present time $t$, the divergence of $q(\bs, \bo)$ from $p(\bs, \bo)$ can be evaluated using 
(\ref{eqn_app:KLHMM}):
\begin{align}
\lefteqn{\KL{q(\bs, \bo)}{p(\bs, \bo)} =} \nonumber \\
& {\sum_{\tau = 1}^T \bq^\top_\tau \ln \bq_\tau  - \sum_{\tau = 1}^{t} \bq^\top_\tau \ln \bA \ \bar  \bo_\tau - \sum_{\tau = 1}^T {\bq}^\top_{\tau - 1} \ln \bB\ {\bq}_{\tau}}. \label{eqn:KLHMM}
\end{align}

The cases $t = 0$ (the beginning of the cycle) and $t = T$ (the end of the cycle) are of particular interest.
In the case $t = 0$, the variables $\bo_1, \ldots, \bo_T$ are yet to be observed, the observation history is empty and
the posteriors $\bq_1, \ldots, \bq_T$  are updated using the prediction update formulas (\ref{eqn:update2}).  
By (\ref{eqn:factorization}), $q(\bo|\bs) = p(\bo|\bs)$ so that
the divergence $\KL{q(\bo|\bs)}{p(\bo|\bs)}$ is 0 and, by the chain rule (\ref{eqn:chain}),
\bE
\label{eqn:future2}
\KL{q(\bs, \bo)}{p(\bs, \bo)} = \KL{q(\bs)}{p(\bs)}.
\eE
This divergence is given explicitly by
\begin{align}
{\sum_{\tau = 1}^T \bq^\top_\tau \ln \bq_\tau  - \sum_{\tau = 1}^T {\bq}^\top_{\tau - 1} \ln \bB\ {\bq}_{\tau}}.  \label{eqn:KLHMM-marginal}
\end{align}

In the case $t = T$, all of the variables $\bo_1, \ldots, \bo_T$ have been observed and the posteriors $\bq_1, \ldots, \bq_T$ are updated using the retrodiction update formulas (\ref{eqn:update1}). 
By (\ref{eqn:factorization}),  $q(\bs, \bo) = q(\bs) \delta_{\bar \bo}(\bo)$
and, by (\ref{eqn:VFE=KL}), the divergence $\KL{q(\bs, \bo)}{p(\bs, \bo)}$ is just
the variational free energy of the observations $\bar \bo_1, \ldots, \bar \bo_T$ calculated with the approximate posterior $q(\bs)$. This divergence is given explicitly by
\begin{align}
 {\sum_{\tau = 1}^T \bq^\top_\tau \ln \bq_\tau  - \sum_{\tau = 1}^T \bq^\top_\tau \ln \bA \ \bar  \bo_\tau - \sum_{\tau = 1}^T {\bq}^\top_{\tau - 1} \ln \bB\ {\bq}_{\tau}}  \label{eqn:KLHMM-VFE}.
\end{align}

More generally, for $t = 0, \ldots, T$, we can split the right-hand side of (\ref{eqn:KLHMM}) into two terms, a contribution from the past an a contribution from the future. That is,
\begin{align}
\KL{q(\bs, \bo)}{p(\bs, \bo)} &= F_{\le t} + F_{>t} \label{eqn:split}
\end{align}
where
\begin{align}
F_{\le t} &= 
 {\sum_{\tau = 1}^t \bq^\top_\tau \ln \bq_\tau  - \sum_{\tau = 1}^t \bq^\top_\tau \ln \bA \ \bar  \bo_\tau - \sum_{\tau = 1}^t {\bq}^\top_{\tau - 1} \ln \bB\ {\bq}_{\tau}} \label{eqn:F1} \nonumber \\
 &= \KL{q(\bs_{\le t}, \bo_{\le t})} {p(\bs_{\le t}, \bo_{\le t})} 
\intertext{which, by (\ref{eqn:VFE=KL}), is just the variational free energy of the observations up to the present time $t$, and}
F_{> t} &= {\sum_{\tau = t + 1}^T \bq^\top_\tau \ln \bq_\tau  - \sum_{\tau = t + 1}^T {\bq}^\top_{\tau - 1} \ln \bB\ {\bq}_{\tau}}  \nonumber \\
&= \KL{q(\bs_{> t})}{p(\bs_{> t})} \label{eqn:G2}
\end{align}
where this divergence is evaluated using $q(\bs_t)$ as the starting distribution at time $t$ rather than $p(\bs_0)$.

\subsection{Planning}
\label{sec:planning}

Consider now an agent having the ability to perform several different types of action, each modelled by a transition probability matrix. 
Suppose that the agent is charged with a task which is characterized by the requirement that the states of world at time $T$ are sampled from a stationary preference distribution $p(\bs_\tau|C)$ 
such as the steady state distribution $p(\bs_\tau|S)$  that we referred to in our discussion of homeostasis in Section \ref{sec:perception}. 
Suppose further that the agent is given a list of the possible action sequences ending at time $T$ that are available to it. These action sequences are called {\em policies} in the active inference literature.

Assume that at a given present time $t = 0, \ldots, T-1$, 
the agent has performed actions $a_1, \ldots, a_t$ and recorded observations $\bar \bo_1, \ldots, \bar \bo_{t}$. Which action should the agent perform at time $t + 1$?

One way to answer this question  would be to start from the observation that the calculations in Section \ref{sec:P/A.1}
 can be
modified straightforwardly to accommodate time-dependent transition probability 
matrices $\bB_{\bpi \tau}$ for each policy $\bpi$ and time $\tau$. So for each policy $\bpi$ that is consistent with the actions that have already been performed up to the present time $t$, we can use the observations $\bar \bo_1, \ldots, \bar \bo_{t}$ 
to calculate variational posteriors $q(\bs_\tau|\bpi)$ for $\tau = 1, \ldots, T$ 
and use these posteriors to evaluate the divergence of $q(\bs_{> t} |\bpi)$ from $p(\bs_{> t}|C)$. The agent would then plan to execute
the policy which minimizes this divergence and this would determine
the action that the agent performs at time $t + 1$, as required.

Another possibility, which is much less computationally expensive but does not seem to have been noticed in the active inference literature, would be to calculate divergences in the forward rather than the reverse direction (as discussed in Section \ref{sec:entropy}). Elsewhere in this paper we use reverse divergences because in those situations the target distribution is intractable and hence has to be approximated. In the case at hand, the target 
distribution, namely $p(\bs_{> t}|C)$, is already factorized and the divergence $\KL{p(\bs_{> t}|C)}{p(\bs_{>t}|\bpi)}$ can be evaluated straightforwardly without having to use the variational approximation
$q(\bs_{> t}|\bpi)$ to $p(\bs_{> t}|\bpi)$. Indeed, if $\bp_C$ is the vector representation of $p(\bs_\tau |C)$ (which we are assuming to be independent of $\tau$), then
\[
\KL{p(\bs_{> t}|C)}{p(\bs_{>t}|\bpi)} =  {\sum_{\tau = t + 1}^T \bp^\top_{C} \ln \bp_{C}  - \sum_{\tau = t + 1}^T {\bp}^\top_{C} \ln \bB_{\bpi \tau} {\bp}_{C}} 
\]
by (\ref{eqn:G2}).

\subsection{Time-Domain Renormalization}
\label{sec:hierarchy}

The approach to planning that we have just outlined has the drawback that the list of policies to be searched can be expected to increase exponentially with the planning horizon, $T$. 
This explosion could be avoided by treating the process of generating a policy as a hidden stochastic process about which the agent makes probabilistic inferences.  
The simplest possibility is to assume that action sequences are generated by a Markov chain so that 
\bE
\label{eqn:Markov}
p(a_0, \ldots, a_T) = p(a_0) \prod_{\tau = 1}^T p(a_\tau|a_{\tau - 1}).
\eE
More generally, we could assume as a generative model for action sequences a second HMM whose states we will denote by $\bsigma_\tau$ so that
\bE
\label{eqn:sigmaHMM}
p(\ba) = p(\bsigma_0) \prod_{\tau = 1}^T p(\bsigma_\tau|\bsigma_{\tau - 1}) p(a_\tau|\bsigma_\tau)
\eE
where $\ba$ is the sequence of actions $a_1, \ldots, a_T$. Under a model of this sort, the search space grows linearly rather than exponentially in $T$.

Note that the two HMMs under consideration can be integrated into a single HMM whose states are triples $(\bsigma_\tau, a_\tau, \bs_\tau)$ and whose transition, emission and starting probability distributions are given by the equations
\begin{align}
p(\bsigma_\tau, a_\tau, \bs_\tau|\bsigma_{\tau - 1}, a_{\tau - 1}, \bs_{\tau - 1}) &= p(\bsigma_\tau|\bsigma_{\tau - 1})p(a_\tau|\bsigma_{\tau})p(\bs_\tau|\bs_{\tau - 1}, a_\tau) \nonumber \\
p(\bo_\tau|\bsigma_\tau, a_\tau, \bs_\tau) &= p(\bo_\tau|\bs_\tau) \nonumber \\
p(\bsigma_0, \bs_0) & = p(\bsigma_0)p(\bs_0). \label{eqn:fold}
\end{align}
Iterating this construction enables a hierarchical structure to be built up. This process is known as 
{\em time domain renormalization} because it can easily accommodate slower clock speeds at higher levels of the hierarchy  \cite{friston_pixels_2024}. 

Although variational inference is traditionally thought of as a method of inferring state occupancies and transitions and active inference as a method 
of inferring actions that are currently underway or yet to be performed, folding
actions into states as in (\ref{eqn:fold}) does away with this distinction. 

Of course the prediction and retrodiction updates (\ref{eqn:update2}) and (\ref{eqn:update1}) would need to be modified if variational posteriors are to be calculated efficiently with a structured HMM of this sort. 
We won't go into details but just point out that a natural mean field factorization to use for this purpose would be to set
\[
q(\bsigma, \ba, \bs, \bo) = q(\bsigma_0)q(\bs_0) \prod_{\tau = 1}^T  q(\bsigma_\tau)p(a_\tau|\bsigma_\tau) q(\bs_\tau, \bo_\tau|a_\tau)
\]
where $q(\bsigma_0) = p(\bsigma_0)$, $q(\bs_0) = p(\bs_0)$ and $q(\bs_\tau, \bo_\tau|a_\tau)$ factorizes in the same way as (\ref{eqn:factorization}).
Updating a full set of variational posteriors $q(\bsigma_\tau, a_\tau, \bs_\tau, \bo_\tau)$ (where $\tau$ ranges from $1$ to $T$) at the present time $t$ gives, in particular, a predictive distribution $q(a_{t+1})$
on the action that will be performed at time $t + 1$. So the agent would decide what to do next by drawing a sample from this distribution.

The reader may wonder what has happened to the stationary preference distribution $p(\bs_\tau|C)$  introduced in the last section. It is generally the case that a finite state Markov chain has a unique steady state distribution \cite{murphy_probabilistic_2023}.
Applied to the integrated HMM (\ref{eqn:fold}), this implies that, regardless of the starting distributions $p(\bsigma_0)$ and $p(\bs_0)$, the triples $(\bsigma_\tau, a_\tau, \bs_\tau)$ 
will eventually be distributed according to a steady state distribution. The same will therefore be true of the states $\bs_\tau$. The latter steady state distribution is determined by the HMM (\ref{eqn:sigmaHMM}), rather than by 
an externally prescribed preference distribution $p(\bs_\tau|C)$. In order to ensure that the steady state and the preference distribution agree, the parameters of the HMM (\ref{eqn:sigmaHMM}) would have to be learned from an appropriate 
training dataset. (We will discuss learning in Section \ref{sec:learning}.)

\section{The Exploration/Exploitation Tradeoff}

\label{sec:info}
As shown in  \cite{hafner_action_2022},  many well known reinforcement learning algorithms can be viewed as solving constrained divergence minimization 
problems. This perspective enables the exploration/exploitation tradeoff in reinforcement learning to be understood in a principled way. In this section, we show how
this analysis plays out in the case of HMMs.

The present time $t$ is fixed throughout this section so we will use the notation $\bo_>$ rather than $\bo_{>t}$ to indicate future observations and similarly for $\bs_>$. Using the factorizations
\begin{align*}
q(\bs_>, \bo_>) &= q(\bo_>|\bs_>)q(\bs_>) \\
p(\bs_>, \bo_>) &= 
p(\bo_>)p(\bs_>|\bo_>),  
\end{align*}
we can write
\begin{eqnarray*}
\lefteqn{\ln q(\bs_>, \bo_>) - \ln p(\bs_>, \bo_>) = } \\
&  (\ln q(\bo_>|\bs_>) - \ln p(\bo_>)) 
+ (\ln q(\bs_>) - \ln p(\bs_>|\bo_>)).
\end{eqnarray*}
So the divergence $\KL{q(\bs_>, \bo_>)}{p(\bs_>, \bo_>)}$ can be written in the form
\bE
\label{eqn:1}
\Esub{q(\bs_>)}{\KL{q(\bo_>|\bs_>)}{p(\bo_>)}} + \Esub{q(\bs_>, \bo_>)}{ \ln q(\bs_>) - \ln p(\bs_>|\bo_>)}.
\eE
Since the divergence of $q(\bs_>|\bo_>)$ from  $p(\bs_>|\bo_>)$ is non-negative, 
\[\Esub{q(\bs_>|\bo_>)}{\ln q(\bs_>|\bo_>)}\ge 
\Esub{q(\bs_>|\bo_>)}{\ln p(\bs_>|\bo_>)} \]
which implies that 
\begin{align}
\label{eqn:2}
\Esub{q(\bs_>, \bo_>)}{\ln q(\bs_>) - \ln p(\bs_>|\bo_>)} &\ge \Esub{q(\bs_>, \bo_>)}{\ln q(\bs_>) - \ln q(\bs_>|\bo_>)} \nonumber \\
&= -\Esub{q(\bs_>)}{\KL{q(\bo_>|\bs_>)}{q(\bo_>)}}  \nonumber \\
&= -{\KL{q(\bs_>,\bo_>)}{q(\bs_>)q(\bo_>)}}.
\end{align}
So combining (\ref{eqn:1}) and (\ref{eqn:2}), we obtain
\begin{eqnarray}
\label{eqn:info}
\lefteqn{\KL{q(\bs_>, \bo_>)}{p(\bs_>, \bo_>)}  \ge} \nonumber \\ 
&  \Esub{q(\bs_>)}{
\KL{q(\bo_>|\bs_>)}{p(\bo_>)}}
- \KL{q(\bs_>, \bo_>)}{q(\bs_>)q(\bo_>)}.
\end{eqnarray}
The expression on the right hand side here is referred to as the {\em free energy of the expected future} in \cite{Millidge2021Whence,tschantz2020reinforcement}. 

The divergence $\KL{q(\bs_>, \bo_>)}{q(\bs_>)q(\bo_>)}$ 
is the {\em mutual information} between $\bs_>$ and $\bo_>$. This is a measure of the degree to which 
$\bs_>$ and $\bo_>$ {\em fail} to be statistically independent. 
In the active inference literature, the mutual information is usually referred to as the
{\em expected information gain} as it can be interpreted as the reduction in uncertainty about $\bs_>$ contingent on observing $\bo_>$ given by the expression
\[
\ent{q(\bs_>)} - \Esub{q(\bo_>)}{\ent{q(\bs_>|\bo_>)}}
\]
which is just another way of writing the mutual information between $\bs_>$ and $\bo_>$. The reason why we did not use a full mean field approximation in (\ref{eqn:factorization}) is to ensure that this reduction in 
uncertainty is non-zero. (If $q(\bs_>, \bo_>)$ factorized as $q(\bs_>)q(\bo_>)$, there would be no reduction in uncertainty.)

Because the mutual information appears in (\ref{eqn:info}) with a negative sign, minimizing the divergence of $q(\bs_>, \bo_>)$ from $p(\bs_>, \bo_>)$ can be expected to increase the information gain, although it
is not guaranteed 
to do so: firstly, because this is an inequality not an equality, and secondly, because the first term on the right-hand side of  (\ref{eqn:info}) depends on $q(\bs_>, \bo_>)$ as well as the second. The first  term                         
can be viewed as a measure of how well  the marginal distribution $p(\bo_>)$ is approximated by $q(\bs_>, \bo_>)$. So it is a {\em pragmatic value} or {\em extrinsic value} in the terminology of active inference.
It can be decreased by acting on the world in a way that ensures that future observations have high probability 
under the marginal distribution $p(\bo_>)$.  But this
sort of action is not feasible if the agent is uncertain as to the current state of its world. In such a situation, the right-hand side of (\ref{eqn:info}) can 
only be minimized by acting in such a way as to maximize the information gain. Thus (\ref{eqn:info}) can be viewed as 
formalizing the tradeoff between exploration and exploitation that an intelligent 
agent needs to be able to make as it navigates its world.

Since $\bo_>$ is yet to be observed,  (\ref{eqn:factorization}) implies that
$q(\bo_>|\bs_>) = p(\bo_>|\bs_>)$ and we can use this to rewrite (\ref{eqn:info}) as follows. 
For the first term on the right-hand side of (\ref{eqn:info}) we have
\begin{align*}
\lefteqn{
\Esub{q(\bs_>)}{\KL{q(\bo_>|\bs_>)}{p(\bo_>)}}  }\\
 & = \Esub{q(\bs_>)}{\KL{p(\bo_>|\bs_>)}{p(\bo_>)}} \\
 &= \Esub{q(\bs_>)} { \Esub{p(\bo_>|\bs_>)}{\ln p(\bo_>|\bs_>) -\ln p(\bo_>)}} \\
&= \Esub{q(\bo_>)}{ -\ln p(\bo_>)} - \Esub{q(\bs_>)}{\ent {p(\bo_>|\bs_>)}}.
\end{align*}
In the active inference literature, the terms extrinsic or pragmatic value are usually reserved for the
cross entropy term on the right-hand side of this equation and the expected entropy term is called the {\em ambiguity}. 
As for the joint divergence on the left-hand side of (\ref{eqn:info}), the assumption that $q(\bo_>|\bs_>) = p(\bo_>|\bs_>)$ implies (by the chain rule (\ref{eqn:chain})) that
\[
\KL{q(\bs_>, \bo_>)}{p(\bs_>, \bo_>)} = \KL{q(\bs_>)}{p(\bs_>)}.
\]
So (\ref{eqn:info}) now takes the form
\begin{eqnarray}
\label{eqn:333}
\lefteqn{\KL{q(\bs_>)}{p(\bs_>)} \ge}\nonumber \\ 
& \Esub{q(\bo_>)}{ -\ln p(\bo_>)} - \Esub{q(\bs_>)}{\ent {p(\bo_>|\bs_>)}} 
\nonumber \\ 
& - \KL{q(\bs_>, \bo_>)}{q(\bs_>)q(\bo_>)}.
\end{eqnarray}
and, since the mutual information between $\bs_>$ and $\bo_>$ can be written in the form 
\[  \ent{q(\bs_>)} - \Esub{q(\bs_>)}{\ent {p(\bo_>|\bs_>)}},
 \]
this simplifies to 
\begin{align}
\KL{q(\bs_>)}{p(\bs_>)} & \ge \Esub{q(\bo_>)}{ -\ln p(\bo_>)}  -\ent{q(\bs_>)}. \label{eqn:simplest}
\end{align}

Transposing the ambiguity term on the right-hand side of (\ref{eqn:333}) to the left-hand side gives
\begin{eqnarray}
\label{eqn:EFE}
\lefteqn{\KL{q(\bs_>)}{p(\bs_>)} + \Esub{q(\bs_>)}{\ent {p(\bo_>|\bs_>)}} \ge} \nonumber \\
& \Esub{q(\bo_>)}{ -\ln p(\bo_>)}  - \KL{q(\bs_>, \bo_>)}{q(\bs_>)q(\bo_>)}.
\end{eqnarray}
In the active inference literature, the expression on the right-hand side of (\ref{eqn:EFE}) is the standard way of defining {\em expected free energy}, although it is suggested  in \cite{Parr2022ActiveInference} that the expression on the left-hand side might be used instead.  

We will discuss expected free energy functionals in Section \ref{sec:whither}. For the moment we note that inspection of  (\ref{eqn:EFE}) shows that
optimizing the right-hand side  tends to decrease the cross-entropy term and increase the mutual information. We say ``tends to'' because both of these terms depend on the approximating distribution $q(\bs_>, \bo_>)$ so that a decrease in the cross entropy may be bought 
at the cost of a decrease in the mutual information (or an increase in the mutual information at the cost of an increase in the cross-entropy). On the other hand
optimizing the right-hand side of (\ref{eqn:simplest}) tends to increase the the entropy $\ent{q(\bs_>)}$ instead of the mutual information.
Optimizing the left-hand side of (\ref{eqn:EFE}) tends to decrease both the divergence term and the ambiguity.

\section{Learning}
\label{sec:learning}

A fully Bayesian approach to the problem of estimating the transition and
emission probability distributions that define a HMM requires that these
probability distributions be treated as random variables. Thus a prior probability
distribution is assigned to the HMM parameters before learning and
a posterior distribution after learning. This posterior distribution can be
updated on an ongoing basis (as more training data becomes available) and
point estimates of the HMM parameters can be calculated from the posterior
as needed. The Dirichlet distribution is the natural choice for a prior on
discrete probability distributions and on discrete HMMs.

\subsection{Dirichlet Priors}
\label{sec:Dirichlet}
The Dirichlet distribution is a multivariate distribution on $K$-tuples $(\mu_1, \ldots, \mu_K)$ whose components are positive and sum to 1. 
It is defined by the probability density
\[
\frac{1}{B(\balpha)} \prod_{k = 1}^K \mu_k^{\alpha_k - 1}.
\]
Here $B(\balpha)$ is the beta function defined by 
\[
B(\balpha) = \frac{\Gamma(\alpha_1) \ldots \Gamma(\alpha_K)}{\Gamma(\alpha_0)}
\]
where $\alpha_0 = \alpha_1 + \ldots + \alpha_K$. The 
$\alpha\/$'s are referred to as concentration parameters
or Dirichlet counts. If $\mu_k$ is interpreted as the probability that an
integer valued random variable takes the value $k$, $\alpha_k$ is usually thought of as
the number of times this event is observed to happen in a virtual training
set although the $\alpha\/$'s are not required to be whole numbers. 
The Dirichlet
distribution has the property that
\bE
\label{eqn:expln}
\E{\ln \mu_k} = \psi(\alpha_k) - \psi(\alpha_0)
\eE
for $k = 1, \ldots, K$. Here $\psi$ is the digamma function (that is, the derivative of $\ln \Gamma$). If $q(\bmu)$ and $p(\bmu)$ are Dirichlet distributions defined by concentration
parameters $\balpha'$ and $\balpha$ respectively then the divergence $\KL{q(\bmu)}{p(\bmu)}$
is given by
\bE
\label{eqn:DirichletDivergence}
\ln B(\balpha) - \ln B(\balpha') + \sum_{k = 1}^K (\alpha'_k - \alpha_k))\left( \psi(\alpha'_k) - \psi(\alpha'_0)\right).
\eE

Recall that a HMM $\bM$  is specified by emission and transition probability
distributions stored as rows of matrices which we denoted by $\bA$ and $\bB$.
We can define a prior on $\bM$ by assigning Dirichlet priors to each of the rows
of the these matrices. It is convenient to write this prior in matrix
form as follows. If $C_\bA$ is the matrix whose entries are the concentration
parameters for the emission probabilities then we can write the prior on $\bA$
in the logarithmic domain as
\[
\ln p(\bA) = {\rm tr}\left( \ln \bA \left( C_\bA - \bOne \right)^\top \right) + \ldots.
\]
where $\bOne$ is the matrix all of whose entries are equal to 1.
Similarly,
\[
\ln p(\bB) = {\rm tr}\left( \ln \bB \left( C_\bB - \bOne \right)^\top \right) + \ldots.
\]
where $C_\bB$ is the matrix whose entries are the concentration parameters for the
transition probabilities. So the prior on $\bM$ has the form 
\bE
\label{eqn:MPrior}
\ln p(\bM) = {\rm tr}\left( \ln \bA \left( C_\bA - \bOne \right)^\top \right) + {\rm tr}\left( \ln \bB \left( C_\bB - \bOne \right)^\top \right)  + \ldots .
\eE

\subsection{Learning as Divergence Minimization}

Suppose we are given a Dirichlet prior $p(\bM)$ as in (\ref{eqn:MPrior})
and a training set consisting of an observation sequence $\bar \bo$ of length $T$. We can calculate a variational posterior on $\bM$ --- which turns out to be a Dirichlet distribution defined by 
another set of concentration parameters --- as follows.

We define a joint prior on triples $(\bM, \bs, \bo)$ by setting
\[
p(\bM, \bs, \bo) = p(\bM) p(\bs, \bo|\bM).
\]
and seek a variational approximation $q(\bM, \bs, \bo)$ to $p(\bM, \bs, \bo)$  of the form
\[
q(\bM, \bs, \bo) = q(\bM) q(\bs) \delta_{\bar \bo}(\bo)
\]
where 
\[
q(\bs) = \prod_{\tau = 0}^T q(\bs_\tau).
\]
As in Section \ref{sec:P/A}, we represent each distribution $q(\bs_\tau)$ by a vector $\bq_\tau$. Following the prescription (\ref{eqn:VBupdate}),
we alternate between updating $q(\bM)$ holding $q(\bs)$ fixed and updating $q(\bs)$ holding $q(\bM)$ fixed. 

Thus we update $q(\bM)$ by 
setting  $q(\bM) \propto \tilde q(\bM)$ where
\begin{align}
\ln \tilde q(\bM) &= \Esub{q(\bs, \bo)}{\ln p(\bM, \bs, \bo)} \nonumber \\
&= \Esub{q(\bs)}{\ln p(\bM, \bs, \bar \bo)} \nonumber \\
&= \ln p(\bM) +  \Esub{q(\bs)}{\ln p(\bs, \bar \bo|\bM)}. \label{eqn:dd}
\end{align}
Recall that by (\ref{eqn:joint}),
\begin{equation}
\ln p(\bs, \bar \bo|\bM) =  {\bs}^\top_0 \ln {\bp}_0 + \sum_{\tau = 1}^T \left(  {\bs}_\tau^\top \ln \bA \  {\bar \bo}_\tau + {\bs}^\top_{\tau - 1} \ln \bB\ {\bs}_{\tau}  \right).
\end{equation}
so the second term on the right hand side of (\ref{eqn:dd}) is equal to
\begin{align*}
 & \sum_{\tau = 1}^T {\bq}_\tau^\top \ln \bA \  {\bar \bo}_\tau + \sum_{\tau  = 1}^T {\bq}^\top_{\tau - 1} \ln \bB \   {\bq}_{\tau}   \\
= & {\rm tr} \left( \ln \bA \ \sum_{\tau = 1}^T \bar \bo_\tau \bq^\top_\tau \right) +  {\rm tr} \left( \ln \bB \ \sum_{\tau = 1}^T \bq_\tau \bq^\top_{\tau - 1}\right).
\end{align*}
Combining this with (\ref{eqn:MPrior}) and collecting terms, (\ref{eqn:dd}) becomes
\begin{align*}
\ln \tilde q(\bM) 
&=  {\rm tr}\left(\ln \bA \  (C'_\bA - \bOne)^\top\right) + {\rm tr}\left(\ln \bB \  (C'_\bB - \bOne)^\top\right) + \ldots
\end{align*}
where
\begin{align}
\label{eqn:posterior}
C'_\bA &= C_\bA + \sum_{\tau = 1}^T \bq_\tau\bar \bo^\top_\tau \nonumber \\
C'_\bB &= C_\bB + \sum_{\tau = 1}^T  \bq_{\tau - 1} \bq_\tau^\top .
\end{align}
So, like $p(\bM)$,  $q(\bM)$ is of the form (\ref{eqn:MPrior}). Thus it too is a Dirichlet distribution.

To update $q(\bs)$ holding $q(\bM)$ fixed, we set $q(\bs_0) = p(\bs_0)$ and for each $\tau = 1, \ldots, T$, we update $q(\bs_\tau)$ by setting $q(\bs_\tau) \propto \tilde q(\bs_\tau)$ where  $\tilde q(\bs_\tau)$ is defined by
\begin{align}
\ln \tilde q(\bs_\tau) &= \Esub{q(\bs_{\backslash \tau})}{\Esub{q(\bM, \bo)}{\ln p(\bM, \bs, \bo)}} \nonumber \\
&= \Esub{q(\bs_{\backslash \tau})}{\Esub{q(\bM)}{\ln p(\bM, \bs, \bar \bo)}}  \nonumber \\
& =  \Esub{q(\bs_{\backslash \tau})}{\Esub{q(\bM)}{\ln p(\bs, \bar \bo | \bM)}} + \ldots 
\end{align}
Recall that by (\ref{eqn:joint}),
\begin{equation}
\label{eqn:joint2}
\ln p(\bs, \bar \bo|\bM) =  {\bs}^\top_0 \ln {\bp}_0 + \sum_{\tau = 1}^T \left(  {\bs}_\tau^\top \ln \bA \  {\bar \bo}_\tau + {\bs}^\top_{\tau - 1} \ln \bB\ {\bs}_{\tau}  \right).
\end{equation}
So 
\begin{equation}
\Esub{q(\bM)}{\ln p(\bs, \bar \bo|\bM)} =  {\bs}^\top_0 \ln {\bp}_0 + \sum_{\tau = 1}^T \left( {\bs}_\tau^\top \left\langle \ln \bA \right\rangle \bar {\bo}_\tau + {\bs}^\top_{\tau - 1} \left\langle \ln \bB \right\rangle  {\bs}_{\tau}  \right)
\end{equation}
where the notation $\langle \ln \bA \rangle$ indicates the expectation of $\ln \bA$ taken with respect to $q(\bM)$ and similarly for $\langle \ln \bB \rangle$. These expectations can be calculated using (\ref{eqn:expln}).
Hence the update formula for $\bq_\tau$ is formally the same as the retrodiction update formula (\ref{eqn:update1}):
\begin{align*}
\ln \bq_\tau
&= \begin{cases}
\ln \bp_0 & \quad (\tau = 0) \\
 \left\langle \ln \bA \right\rangle \ \bar \bo_\tau + \left\langle  \ln  \bB \right\rangle^\top \bq_{\tau-1} + \left\langle \ln \bB \right\rangle \ \bq_{\tau + 1}   + \ldots& \quad (\tau = 1, \ldots, T- 1)\\
 \left\langle \ln \bA \right\rangle \ \bar \bo_T + \left\langle  \ln \bB \right\rangle^\top \bq_{T-1}    + \ldots & \quad (\tau = T).
\end{cases}
\end{align*}

It remains to evaluate the joint divergence  $\KL{q(\bM, \bs, \bo)}{p(\bM, \bs, \bo)}$. By the chain rule (\ref{eqn:chain}), we can  write this as 
\[
\KL{q(\bM)}{p(\bM)} + \Esub{q(\bM)}{\KL{q(\bs,\bo)}{p(\bs, \bo|\bM)}}.
\]
The first term here is given by (\ref{eqn:DirichletDivergence}). For the second, since $\bo$ is observed to take the value $\bar \bo$, the divergence $\KL{q(\bs, \bo | \bM)}{p(\bs, \bo | \bM)}$ is given by the variational free energy formula (\ref{eqn:KLHMM-VFE}):
\begin{align*}
\lefteqn{\KL{q(\bs, \bo | \bM)}{p(\bs, \bo | \bM)}} \\
&= {\sum_{\tau = 1}^T \bq^\top_\tau \ln \bq_\tau   - \sum_{\tau = 1}^T \bq^\top_\tau \ln \bA \ \bar \bo_\tau
- \sum_{\tau = 1}^T {\bq}^\top_{\tau - 1} \ln \bB\ {\bq}_{\tau}}
\end{align*}
so that 
\begin{align*}
\lefteqn{
\Esub{q(\bM)}{\KL{q(\bs, \bo)}{p(\bs, \bo|\bM)}}} \\
& =  \sum_{\tau = 1}^T \bq^\top_\tau \ln \bq_\tau  - \sum_{\tau = 1}^T {\bq}_\tau^\top \langle \ln \bA   \rangle  \bar {\bo}_\tau - \sum_{\tau  =1}^T {\bq}^\top_{\tau - 1} \langle \ln \bB   \rangle  {\bq}_{\tau}.  
\end{align*}

Extending these derivations to accommodate the structured state spaces discussed in Section \ref{sec:hierarchy} is straightforward. 
Consider the matrix $\sum_{\tau = 1}^T  \bq_{\tau - 1} \bq_\tau^\top$ appearing the second line of (\ref{eqn:posterior}). For each transition from one state to another,
there is a corresponding entry in this matrix whose value is the expected number of times the transition occurs in the training data. If states are decomposed as in (\ref{eqn:fold}) or into more complex hierarchies, then all that is required is to accumulate 
matrices of expected transition counts for each level in the hierarchy.

\section{Whither Expected Free Energy?}
\label{sec:whither}

Recall that, given observations that have already been made, variational free energy is an approximation to the surprisal of these observations.  Evaluating it requires a generative model $p(\bs, \bo)$ (such as a HMM), the observations $\bo$ themselves, and an approximation to the 
posterior distribution  $p(\bs|\bo)$. 

If we can calculate an approximate posterior $q(\bs_>|\bo_>)$ for observations $\bo_>$ that have yet to be made, 
and in addition we have an approximate marginal distribution $q(\bo_>)$ on these observations, then we
can calculate an expected value for the variational free energy of future observations. This expected value is traditionally denoted by $G$.  How should $G$ be evaluated?

\subsection{Expected Free Energy Functionals}
\label{sec:functionals}

Note that, taken together, $q(\bo_>)$ and $q(\bs_>|\bo_>)$ define a joint distribution
$q(\bs_>, \bo_>)$. 
We assume that this joint distribution satisfies the condition that  
$q(\bo_>|\bs_>) = p(\bo_>|\bs_>)$ in order to ensure that the mutual information between $\bs_>$ and $\bo_>$ is non zero (as explained in  Section \ref{sec:info}).

Although this assumption seems to be needlessly restrictive (as seems to have been first pointed out in \cite{Millidge2021Whence}), the approximation 
\bE
q(\bs_>|\bo_>) \approx q(\bs_>) \label{eqn:approx1}
\eE
enables $G$ to be evaluated easily.
The variational free energy for $\bo_>$ is \[ 
\Esub{q(\bs_>)}{ - \ln p(\bs_>,\bo_>)} - \ent {q(\bs_>)}
\]
so, taking the expectation with respect to $q(\bo_>)$,
\begin{align}
G{} &= \Esub{q(\bo_>)q(\bs_>)}{ - \ln p(\bs_>,\bo_>)} - \ent {q(\bs_>)} \nonumber \\
&= - \Esub{q(\bs_>)}{\ln p(\bs_>)} - \ent {q(\bs_>)} +  \Esub{q(\bs_>)q(\bo_>)}{ - \ln p(\bo_>|\bs_>)}  \nonumber \\
&= \KL{q(\bs_>)}{p(\bs_>)} + \Esub{q(\bs_>)}{\ent{p(\bo_>|\bs_>)}} \label{eqn:LHS}
\end{align}
which is just the left-hand side of (\ref{eqn:EFE}). However, it is the expression on the right-hand side of (\ref{eqn:EFE}) that is taken as standard definition of expected free energy in the active inference 
literature. This can be derived by appealing to an additional approximation, namely
\bE
p(\bs_>, \bo_>) \approx q(\bs_>|\bo_>) p(\bo_>) \label{eqn:approx2}.
\eE
Using both of these approximations, we have
\begin{align}
G{}  &= \Esub{q(\bs_>, \bo_>)}{ - \ln p(\bs_>,\bo_>)} 
-\Esub{q(\bo_>)}{\ent{q(\bs_>|\bo_>)}} \nonumber \\
& \approx
\Esub{q(\bs_>, \bo_>)}{ -\ln q(\bs_>|\bo_>) - \ln p(\bo_>)}  -\ent{q(\bs_>)} \nonumber \\
& =\Esub{q(\bs_>,\bo_>)}{ - \ln q(\bs_>,\bo_>) +
q(\bs_>) + q(\bo_>) } + \Esub{q(\bo_>)}{ -\ln p(\bo_>)} \nonumber \\
&= -
\KL{q(\bs_>, \bo_>)}{q(\bs_>)q(\bo_>)} +
\Esub{q(\bo_>)}{ - \ln p(\bo_>)} \label{eqn:standardEFE}
\end{align}
which is just the right-hand side of (\ref{eqn:EFE}), as required.  

What happens if we proceed without making either of these approximations? The variational free energy for $\bo_>$ is
\[ 
\Esub{q(\bs_>|\bo_>)}{ - \ln p(\bs_>,\bo_>)} - \ent {q(\bs_>|\bo_>)}
\]
so, taking the expectation with respect to $q(\bo_>)$,  we obtain the following expression for the expected free energy:
\bE
\label{eqn:G}
G{} = \Esub{q(\bs_>, \bo_>)}{ - \ln p(\bs_>,\bo_>)} 
-\Esub{q(\bo_>)}{\ent{q(\bs_>|\bo_>)}}.
\eE
Since
\[
\ent{q(\bs_>, \bo_>)} = \ent{q(\bo_>)} + \Esub{q(\bo_>)}{\ent{q(\bs_>|\bo_>)}}, 
\]
we can rewrite this as
\begin{align}
	G{} &= \Esub{q(\bs_>, \bo_>)}{- \ln p(\bs_>,\bo_>)} - \ent{q(\bs_>, \bo_>)} + \ent{q(\bo_>)} \nonumber \\
      	&= \KL{q(\bs_>, \bo_>)}{p(\bs_>, \bo_>)} + \ent{q(\bo_>)} \nonumber \\
	&= \KL{q(\bs_>)}{p(\bs_>)} + \ent{q(\bo_>)} \label{eqn:xxxx}
\end{align}
since we are assuming that $q(\bo_>|\bs_>) = p(\bo_>|\bs_>)$. 
So if the expected free energy $G$ is defined in this way, the marginal divergence $\KL{q(\bs_>)}{p(\bs_>)}$ 
and $G$ are very closely related:
\bE
\label{eqn:GKL}
\KL{q(\bs_>)}{p(\bs_>)} = G  - \ent{q(\bo_>)}.
\eE
Comparing this with the general relationship (\ref{eqn:KLCE})
\[
\KL{q(\bx)}{p(\bx)} = \Esub{q(\bx)}{-\ln p(\bx)} - \ent{q(\bx)},
\]
which we discussed in Section \ref{sec:entropy}, we see that the entropy term $ \ent{q(\bo_>)}$ in (\ref{eqn:GKL}) can be interpreted as a regularizer which can serve to prevent
overfitting when the divergence $\KL{q(\bs_>)}{p(\bs_>)}$ is used
as a criterion for approximating $p(\bs_>)$ by $q(\bs_>)$. 

\subsection{Beliefs about Policies}
\label{sec:policy}

Expected free energy is usually thought of as the objective function which an agent needs to optimize in order to make probabilistic inferences about actions which are currently underway or yet to be performed, or more generally, to update beliefs about policies.

Suppose that, at present time $t$, an agent has a prior distribution $p(\bpi)$ over policies where each policy is a sequence of actions $a_1, \ldots, a_T$ and that observations $\bar \bo_{\le t}$ have been recorded. According to equation (B.9) of \cite{Parr2022ActiveInference}, the agent's posterior belief about policies is given by
\[
q(\bpi) = \sigma\left(\ln p(\bpi) - F(\bpi) - G(\bpi) \right)
\]
where $\sigma$ is the softmax function and, for each policy $\bpi$, $F(\bpi)$ is the variational free energy associated with the observation history, and $G(\bpi)$ 
is the expected free energy associated with future observations. 

To see how such an equation might be derived, let us define a joint distribution on triples $(\bpi, \bs, \bo)$ by setting
\bE
\label{eqn:jointPolicy}
p(\bpi, \bs, \bo) = p(\bpi)p(\bs, \bo|\bpi),
\eE
We can obtain an approximate posterior distribution $q(\bpi)$ on $\bpi$ by assuming a variational approximation to this joint distribution of the form  
$q(\bpi) q(\bs, \bo)$ where $q(\bs, \bo)$ factorizes as in (\ref{eqn:factorization}). If $q(\bs, \bo)$ is held fixed then, by (\ref{eqn:VBupdate}), the update formula for $q(\bpi)$ is $q(\bpi) \propto \tilde q(\bpi)$ where
\begin{align*}
\ln \tilde q(\bpi) &= \ln p(\bpi) + \Esub{q(\bs, \bo)}{\ln p(\bs, \bo|\bpi) } \\
& = \ln p(\bpi) - \KL{q(\bs, \bo)}{ p(\bs, \bo|\bpi)} + \ldots
\end{align*}
since  by (\ref{eqn:KLCE}),
\[
\KL{q(\bs, \bo)}{ p(\bs, \bo|\bpi)} = \Esub{q(\bs, \bo)}{- \ln p(\bs, \bo|\bpi) } - \ent{q(\bs, \bo)}
\] 
and the entropy term $\ent{q(\bs, \bo)}$ is independent of $\bpi$. 
As in (\ref{eqn:split}), we can write
\begin{align}
\KL{q(\bs, \bo)}{p(\bs, \bo|\bpi)}  & = F_{\le }(\bpi) + F_>(\bpi) \nonumber
\intertext{where}
F_\le (\bpi) &= \KL{q(\bs_{\le }, \bo_{\le })} {p(\bs_{\le }, \bo_{\le }|\bpi)}  \nonumber
\intertext{and}
F_>(\bpi) &=  \KL{q(\bs_>)}{p(\bs_>|\bpi)} \label{eqn:Ffuture}
\end{align}   
where this divergence is evaluated using $q(\bs_t)$ 
as the starting distribution at the present time $t$. 
So the update formula for $q(\bpi)$ is
\bE
\label{eqn:posteriorPolicy}
q(\bpi) = \sigma\left(\ln p(\bpi) - F_\le(\bpi) - F_>(\bpi) \right)
\eE
which yields equation (B.9) of \cite{Parr2022ActiveInference} if $G(\bpi)$ is taken to be $F_>(\bpi)$. 

Alternatively, we could take 
\bE
\label{eqn:GG}
G(\bpi) = \Esub{q(\bs_{>}, \bo_>)} {- \ln p(\bs_>, \bo_> |\bpi)} - \Esub{q(\bo_{>})}{\ent{q(\bs_>|\bo_>)}}
\eE
since, as in the derivation of (\ref{eqn:GKL}) from (\ref{eqn:G}), 
\bE
F_>(\bpi) = G(\bpi)  - \ent{q(\bo_>)}.
\eE
The fact that the entropy term on the right-hand side here is independent of $\bpi$ implies that the values returned 
by the softmax function would be unaffected by substituting the right-hand side of (\ref{eqn:GG}) for $F_>(\bpi)$ in  (\ref{eqn:posteriorPolicy}).

The expression for $G(\bpi)$ in (\ref{eqn:GG}) has the same form as the free energy functional in  (\ref{eqn:G}) which we derived without appealing to either of the approximations (\ref{eqn:approx1}) and (\ref{eqn:approx2}).  Note however that the argument we have just given depends
on calculating $F_\le(\bpi)$ and $F_>(\bpi)$ with the  distribution $q(\bs, \bo)$ postulated by the variational approximation to the joint distribution defined by (\ref{eqn:jointPolicy}). This is {\em not} the same as the policy dependent mean field approximation to $p(\bs, \bo|\bpi)$
that would be obtained by applying the prediction and retrodiction update formulas (\ref{eqn:update2}) and (\ref{eqn:update1}) to each policy individually. So this way of defining variational and expected free energy for policies
depends on how the distribution $q(\bs, \bo)$ is estimated. The correct estimation procedure is to update $q(\bs, \bo)$ and $q(\bpi)$ alternately in the usual way. 

We have already derived the update formula for $q(\bpi)$ holding  $q(\bs, \bo)$ fixed so it only remains to derive the update formula for $q(\bs, \bo)$ holding $q(\bpi)$ fixed. By (\ref{eqn:VBupdate}), $q(\bs, \bo) \propto \tilde q(\bs, \bo)$ where
\begin{align*}
\ln \tilde q(\bs, \bo) &= \Esub{q(\bpi)}{\ln p(\bpi, \bs, \bo))}.
\end{align*}
By (\ref{eqn:joint}),
\[
\ln p(\bs, \bo|\bpi) =  {\bs}^\top_0 \ln {\bp}_0  + \sum_{\tau = 1}^T \left( {\bs}_\tau^\top \ln \bA  \ {\bo}_\tau  +  {\bs}^\top_{\tau - 1} \ln \bB_{\bpi \tau}  {\bs}_{\tau} \right) .
\]
So
\[
\Esub{q(\bpi)}{\ln p(\bs, \bo|\bpi)} =  {\bs}^\top_0 \ln {\bp}_0  + \sum_{\tau = 1}^T \left( {\bs}_\tau^\top \ln \bA  \ {\bo}_\tau  +  {\bs}^\top_{\tau - 1}  \ln \bB_{\tau}   {\bs}_{\tau} \right) 
\]
where $\bB_\tau$ is defined by
\[
\ln \bB_{\tau}  = \sum_{\bpi} 
q(\bpi) \ln \bB_{\bpi \tau} 
\]
So if the matrices $\bB_\tau$ are defined in this way, the update formulas for $q(\bs)$ have the same form as the prediction and retrodiction update formulas (\ref{eqn:update2}) and (\ref{eqn:update1}).

\subsection{Discussion}

Although it is not the expected value of a variational free energy functional,  
the expression on the right-hand side of (\ref{eqn:EFE}) is generally taken as the definition of expected free energy in the active inference literature thanks to its interpretation in terms of pragmatic value and expected information gain. We have seen how this is usually justified
by appealing to the approximation  (\ref{eqn:approx2}). (This approximation is stated explicitly in 
\cite{Sajid2021Active, Smith2022}. 
Other survey articles and  tutorials  take the right-hand side
of (\ref{eqn:EFE}) as the definition of expected free energy without attempting to justify it \cite{oostrum_concise_2025,DaCosta2020_DiscreteActiveInference,zhang2024_overview_fep}.)  

Inspecting the derivation (\ref{eqn:standardEFE}) shows that it also invokes the approximation (\ref{eqn:approx1}).
If it is assumed that $q(\bs_>|\bo_>) = q(\bs_>)$ and that $p(\bs_>, \bo_>) = q(\bs_>|\bo_>)p(\bo_>)$, 
then $\bs_>$ and $\bo_>$ are statistically independent under both the the target distribution
$p(\bs_>, \bo_>)$ and the approximating distribution $q(\bs_>, \bo_>)$ so that the mutual information between $\bs_>$ and $\bo_>$ is zero. This is clearly problematic.

If neither approximation is used then, as we saw in (\ref{eqn:GKL}), the expression obtained for the expected free energy only differs from the divergence $\KL{q(\bs_>)}{p(\bs_>)}$ by an entropy regularizer, $\ent{q(\bo_>)}$. Whether this sort of regularization is practically useful can only be determined by running experiments but is easy to imagine that 
a guardrail against making over confident predictions about future events could prove useful for an agent engaged in planning over extended time scales.
As we discussed in Section \ref{sec:MFA}, the variational updates (\ref{eqn:VBupdate}) are designed to optimize divergences (such as $\KL{q(\bs_>)}{p(\bs_>)}$) 
rather than functionals of the form (\ref{eqn:xxxx}).  So if an expected free energy functional such as $G$ (or, for that mattter, any criterion other than a reverse KL divergence)
were to be used as the objective function for active inference, another algorithm would need to be developed in order to optimize it.

Expected free energy is usually thought of as furnishing an objective function for variational inference about actions and policies as distinct from states and transitions as in (\ref{eqn:posteriorPolicy}). 
However the distinction between states and actions is not absolute as it hinges on the way states are defined. (For example, this distinction is mooted by folding actions into states as in Section \ref{sec:hierarchy}.) Although we had no need to appeal to an equation like (B.9) of \cite{Parr2022ActiveInference}  in the body of this paper, we showed how such an equation can be derived
by defining  variational and expected free energies for policies in terms of a mean field approximation to the joint distribution defined by (\ref{eqn:jointPolicy}). This entails defining the expected free energy associated with a policy by 
(\ref{eqn:Ffuture}) or by (\ref{eqn:GG}) rather than by the right-hand side of (\ref{eqn:EFE}) as is usually done in the active inference literature.

\section{Conclusion}

This work was motivated by a desire to find an objective function for active inference and develop a variational algorithm to optimize it which has the property that the objective function
is guaranteed to decrease on every belief update. As we saw in Section \ref{sec:MFA}, in order to satisfy this condition the objective function must be expressible as a Kullback-Leibler divergence. 
Variational free energy has this property by (\ref{eqn:VFE=KL}) but expected free energy, regardless of how it is defined, does not.

We showed that if an agent models its world by a Hidden Markov Model $p(\bs, \bo)$, it can solve the problems 
of perceptual and active inference in a unified way by using variational inference to optimize a divergence of the form $\KL{q(\bs, \bo)}{p(\bs, \bo)}$ where $q(\bs, \bo)$ is a 
suitably factorized approximation to $p(\bs, \bo)$.
Assuming that at a given time present time $t$ the variables $\bo_{\le t}$ have been observed to take
the values $\bar \bo_{\le t}$, the constraint that we imposed is
\begin{align*}
q(\bs_\tau, \bo_\tau) &= \begin{cases}
q(\bs_\tau) \delta_{\bar \bo_\tau}(\bo_\tau) &\quad (\tau \le t)\\
q(\bs_\tau) p(\bo_\tau|\bs_\tau) &\quad (\tau > t)
\end{cases}
\end{align*}
as in (\ref{eqn:factorization}). The update formulas needed to optimize $q(\bs, \bo)$ subject to this constraint are given in the appendix. 
These formulas are guaranteed to decrease the divergence (given explicitly by (\ref{eqn_app:KLHMM})) on each iteration and their  
derivation makes no appeal to the Free Energy Principle.

We teased out the relationship between this divergence criterion and the notion of expected free energy in Section \ref{sec:whither}. 
Equation (\ref{eqn:split}) shows that the divergence from $q(\bs, \bo)$ to $p(\bs, \bo)$ can be decomposed as 
\begin{align*}
  \KL{q(\bs_{\le t}, \bo_{\le t})} {p(\bs_{\le t}, \bo_{\le t})} + \KL{q(\bs_{> t})}{p(\bs_{> t})}
\end{align*}
where the divergence in the second term is calculated with the starting distribution $q(\bs_t)$ at the present time $t$. By (\ref{eqn:VFE=KL}),  
the first term here is the variational free energy of the observations up to time $t$ calculated with the variational posterior $q(\bs_{\le t})$. The second term can be written in the form
\[
\KL{q(\bs_{> t})}{p(\bs_{> t})} = G - \ent{q(\bo_{> t})}
\]
where $G$ is the expected free energy defined by (\ref{eqn:G})
\[
G = \Esub{q(\bs_{> t}, \bo_{> t})}{ - \ln p(\bs_{> t},\bo_{> t})} 
-\Esub{q(\bo_{> t})}{\ent{q(\bs_{> t}|\bo_{> t})}}.
\]
This was derived without appealing to either of the approximations (\ref{eqn:approx1}) or (\ref{eqn:approx2}). So our approach 
is faithful to the spirit of the Free Energy Principle even as it dispenses with the need to appeal to it. The entropy regularizer $\ent{q(\bo_{> t})}$ which differentiates the divergence $\KL{q(\bs_{> t})}{p(\bs_{> t})}$ from the expected free energy $G$
can be thought of as penalizing over confident predictions of future events or as imposing a soft constraint on the degree to which an agent can look ahead 
in planning its future.

In addition to providing a solution to the problem of calculating approximate predictive distributions given a starting distribution and an observation history, we have seen how standard mean field methods can be brought to bear on the problems of learning HMM parameters (Section \ref{sec:learning}) and updating beliefs about policies (Section \ref{sec:policy}). All of these applications follow the same pattern: in each case, the variational update formulas are determined by the joint distribution of the random variables of interest and the way the approximating distribution is required to factorize. The practitioner who cleaves to orthodox mean field methods stands to benefit from the monotone convergence guarantee discussed in Section \ref{sec:MFA}
but he does not have the latitude claimed by the Free Energy Principle  to choose variational update formulas at will.

\cleardoublepage

\addcontentsline{toc}{section}{References}
\bibliographystyle{ieeetr}
\bibliography{references.bib}
\nocite{*}

\begin{appendices}

\section{Posterior Update Formulas}

The simplest way to minimize the divergence from $q(\bs, \bo)$ to $p(\bs, \bo)$ is to invoke the decomposition (\ref{eqn:split}),  
\begin{align*}
\KL{q(\bs, \bo)}{p(\bs, \bo)} = \KL{q(\bs_{\le t}, \bo_{\le t})} {p(\bs_{\le t}, \bo_{\le t})} + \KL{q(\bs_{> t})}{p(\bs_{> t})},
\end{align*}
(where the divergence in the second term is calculated with the starting distribution $q(\bs_t)$ at the present time $t$) and use the mean field
approximation to update the posteriors $q(\bs_{\le t})$ and $q(\bs_{>t})$ separately.
        
\subsection{Predicting the Future}
\label{predicting}

Fix a time $\tau$ in the range $t + 1, \ldots, T$. We derive the formula for updating $q(\bs_\tau)$ holding fixed the other posteriors $q(\bs_{\tau'})$ (where $\tau'$ takes values in the range $t + 1, \ldots, T$ other than $\tau$).

By (\ref{eqn:VBupdate}), to update $q(\bs_\tau)$ we need to evaluate $\tilde q(\bs_\tau)$ defined by
\begin{align}
\ln \tilde q(\bs_\tau) &= \Esub{q(\bs_{\backslash \tau})}{\ln p(\bs)}. \label{eqn_app:ptilde}
\end{align}
Note that by (\ref{eqn:bb}),
\begin{align}
\Esub{q(\bs_{\backslash \tau})}{\bs_{\tau'}} &= \begin{cases}
\bs_{\tau} & \text{if $\tau' = \tau$} \\
\bq_{\tau'} & \text{otherwise}
\end{cases} \nonumber \\
\intertext{and}
\Esub{q(\bs_{\backslash \tau})}{\bs_{\tau' - 1} \ln \bB \ \bs_{\tau'}} & = \begin{cases}
\bs^\top_{\tau} \ln \bB \ \bq_{\tau + 1} & \text {if $\tau' = \tau - 1$}  \\
\bq^\top_{\tau - 1} \ln \bB \ \bs_\tau & \text {if $\tau' = \tau$}  \\
\bq^\top_{\tau - 1} \ln \bB \ \bq_\tau & \text {otherwise} 
\end{cases}  \nonumber \\
\intertext{so that}
\Esub{q(\bs_{\backslash \tau})}{\ln p(\bs)} &= \Esub{q(\bs_{\backslash \tau})}{{\bs}^\top_0 \ln {\bp}_0 + \sum_{\tau' = 1}^T {\bs}^\top_{\tau' - 1} \ln \bB\ {\bs}_{\tau'} } \nonumber \\
&= \begin{cases}
\bs^\top_\tau \left( \ln \bB^\top \bq_{\tau-1} + \ln \bB \ \bq_{\tau + 1} \right)  + \ldots \quad & (\tau  = 1, \ldots, T - 1) \\
\bs^\top_T \ln \bB^\top \bq_{T-1}  + \ldots \quad & (\tau = T).
\end{cases} \label{eqn_app:first term}
\end{align}

%
Thus by (\ref{eqn:VBupdate}), the update formula for $\bq_\tau$ is
\begin{align}
\label{eqn_app:update2}
\ln \bq_\tau &= 
\begin{cases}
 \ln \bB^\top \bq_{\tau-1} + \ln \bB \ \bq_{\tau + 1}   + \ldots& \quad  (\tau  = 1, \ldots, T - 1) \\
 \ln \bB^\top \bq_{T-1}  + \ldots & \quad (\tau = T).
\end{cases}
\end{align}
if $\tau > t$. 

\subsection{Retrodicting the Past}
\label{retrodicting}

Now fix a present time $t$ and a time $\tau$ in the range $1, \ldots, t$. We derive the formula for updating $q(\bs_\tau)$ holding fixed the other posteriors $q(\bs_{\tau'})$ (where $\tau'$ takes any value in the range $1, \ldots, t$ other than $\tau$).

We need to evaluate $\tilde q(\bs_\tau)$ where
\begin{align}
\ln \tilde q(\bs_\tau) &= \Esub{q(\bs_{\backslash \tau})}{\ln p(\bs)} + \Esub{q(\bs_{\backslash \tau}, \bo)}{\ln p(\bo|\bs)}. \label{eqn_app:ptilde2}
\end{align}
The first term on the right-hand side of (\ref{eqn_app:ptilde2}) is handled in the same way as 
(\ref{eqn_app:first term}) so that
\begin{align}
\Esub{q(\bs_{\backslash \tau})}{\ln p(\bs)} &= \Esub{q(\bs_{\backslash \tau})}{{\bs}^\top_0 \ln {\bp}_0 + \sum_{\tau' = 1}^T {\bs}^\top_{\tau' - 1} \ln \bB\ {\bs}_{\tau'} } \nonumber \\
&= \begin{cases}
\bs^\top_\tau \left( \ln \bB^\top \bq_{\tau-1} + \ln \bB \ \bq_{\tau + 1} \right)  + \ldots& \quad  (\tau  = 1, \ldots, T - 1) \\
\bs^\top_T \ln \bB^\top \bq_{T-1}  + \ldots & \quad (\tau = T).
\end{cases} \label{eqn_app:A}
\end{align}
To evaluate the second term on the right-hand side of (\ref{eqn_app:ptilde2}), note that since we are assuming that $\tau \le t$, (\ref{eqn:factorization}) implies that
\begin{align}
\Esub{q(\bs_{\backslash \tau}, \bo)} {\ln p(\bo_{\tau'}|\bs_{\tau'})} &= 
\begin{cases}
 \bs^\top_\tau \ln \bA \bar  \bo_\tau  & (\tau' = \tau) \\
\bq^\top_{\tau'} \ln \bA \bar  \bo_{\tau'} & (\tau' \neq \tau)
\end{cases} \nonumber \\
\intertext{so that}
\Esub{q(\bs_{\backslash \tau}, \bo)}{\ln p(\bo|\bs)}
&= \sum_{\tau'= 1}^T \Esub{q(\bs_{\backslash \tau}, \bo)}{\ln p(\bo_{\tau'}|\bs_{\tau'})}  \nonumber \\
& = 
\bs^\top_\tau \ln \bA \ \bar \bo_\tau + \ldots . 
 \label{eqn_app:B}
\end{align}
Combining (\ref{eqn_app:A}) and (\ref{eqn_app:B}), 
\begin{align*}
\ln \tilde q(\bs_\tau)
&= \begin{cases}
\bs^\top_\tau \left( \ln \bA \ \bar \bo_\tau  + \ln \bB^\top \bq_{\tau-1} + \ln \bB \ \bq_{\tau + 1} \right)   + \ldots& \quad (\tau =1, \ldots,  T-1) \\
\bs^\top_T \left(\ln \bA \ \bar \bo_T + \ln \bB^\top \bq_{T-1}  \right) + \ldots & \quad (\tau = T)
\end{cases}
\end{align*}
and the update formula for $\bq_\tau$ is
\begin{align}
\ln \bq_\tau
&= \begin{cases}
 \ln \bA \  \bar \bo_\tau   + \ln \bB^\top \bq_{\tau-1} + \ln \bB \ \bq_{\tau + 1}  \ldots& \quad  (\tau  = 1, \ldots, T - 1) \\
\ln \bA \  \bar \bo_T + \ln \bB^\top \bq_{T-1}  +  \ldots & \quad (\tau = T).
\end{cases} \label{eqn_app:update1}
\end{align}
if $\tau \le t$. 

\subsection{Evaluating the Divergence}
After updating all of the posteriors at the present time $t$, we can evaluate the divergence
$\KL{q(\bs, \bo)}{p(\bs, \bo)}$ by rewriting it as
\begin{align}
\lefteqn{\KL{q(\bs, \bo)}{p(\bs, \bo)}}  \nonumber  \\
&= \Esub{q(\bs, \bo)}{\ln q(\bs, \bo) - \ln p(\bs, \bo)} \nonumber \\
&= \Esub{q(\bs)}{\ln q(\bs) - \ln p(\bs)} + \Esub{q(\bs, \bo)}{\ln q(\bo|\bs) - \ln p(\bo|\bs)} \label{eqn_app:divjoint}
\end{align}
and evaluating these expectations separately. Thus
\begin{align*}
\lefteqn{ \Esub{q(\bs)}{\ln q(\bs) - \ln p(\bs)}} \nonumber \\
 &=
\Esub{q(\bs)}{\sum_{\tau = 0}^T \bs^\top_\tau \ln \bq_\tau -{\bs}^\top_0 \ln {\bp}_0 - \sum_{\tau = 1}^T {\bs}^\top_{\tau - 1} \ln \bB\ {\bs}_{\tau}} \nonumber \\
&= {\sum_{\tau = 0}^T \bq^\top_\tau \ln \bq_\tau -{\bq}^\top_0 \ln {\bp}_0 - \sum_{\tau = 1}^T {\bq}^\top_{\tau - 1} \ln \bB\ {\bq}_{\tau}}  \\
&= {\sum_{\tau = 1}^T \bq^\top_\tau \ln \bq_\tau  - \sum_{\tau = 1}^T {\bq}^\top_{\tau - 1} \ln \bB\ {\bq}_{\tau}}  \\
\intertext{since we are assuming that $\bq_0 = \bp_0$. And}
\lefteqn{\Esub{q(\bs, \bo)}{\ln q(\bo|\bs) - \ln p(\bo|\bs)}} \nonumber \\
 &= \sum_{\tau = 1}^{t} \Esub{q(\bs_\tau, \bo_\tau)}{\ln q(\bo_\tau|\bs_\tau) - \ln p(\bo_\tau|\bs_\tau)} \nonumber \\
&= - \sum_{\tau = 1}^{t} \Esub{q(\bs_\tau)}{ \ln p( \bar \bo_\tau|\bs_\tau)} \quad \text{by (\ref{eqn:factorization}) }\nonumber \\
&= - \sum_{\tau = 1}^{t} \Esub{q(\bs_\tau)}{ \bs^\top_\tau \ln \bA \ \bar \bo_\tau} \nonumber \\
&= - \sum_{\tau = 1}^{t} \bq^\top_\tau \ln \bA \ \bar \bo_\tau \quad \text{by (\ref{eqn:bb})}.
\end{align*}
So (\ref{eqn_app:divjoint}) gives
\begin{align}
\lefteqn{\KL{q(\bs, \bo)}{p(\bs, \bo)}} \nonumber \\
&= {\sum_{\tau = 1}^T \bq^\top_\tau \ln \bq_\tau  - \sum_{\tau = 1}^{t} \bq^\top_\tau \ln \bA \ \bar  \bo_\tau - \sum_{\tau = 1}^T {\bq}^\top_{\tau - 1} \ln \bB\ {\bq}_{\tau}}.  \label{eqn_app:KLHMM}
\end{align}

\end{appendices}

\end{document}